\documentclass{article}

\usepackage{microtype}
\usepackage{graphicx}
\usepackage{booktabs}
\usepackage{multirow}
\usepackage{makecell}
\usepackage{amsmath}
\usepackage{amssymb}
\usepackage{bm}
\usepackage[table]{xcolor}
\usepackage{enumitem}
\usepackage{tikz}
\usetikzlibrary{arrows.meta,positioning,fit,calc,shapes.misc}
\usepackage{hyperref}
\usepackage{xurl}
\usepackage{fvextra}

\usepackage[accepted]{mlsys2025}
\makeatletter
\renewcommand{\mlsys@appearing}{Preprint.}
\renewcommand{\printAffiliationsAndNotice}[1]{%
\stepcounter{@affiliationcounter}%
{\let\thefootnote\relax\footnotetext{\hspace*{-\footnotesep}#1%
\forloop{@affilnum}{1}{\value{@affilnum} < \value{@affiliationcounter}}{
\textsuperscript{\arabic{@affilnum}}\csname @affilname\the@affilnum\endcsname}.

\ \\
\Notice@String
}}}
\makeatother

\newcommand{\sys}{\textsc{DMR}}
\newcommand{\tiers}{\mathcal{B}}
\newcommand{\btop}{b_{\mathrm{H}}}
\newcommand{\blow}{b_{\mathrm{L}}}
\newcommand{\hid}{\bm{h}}

\newcommand{\para}[1]{\noindent\textbf{#1}\hspace{0.5em}}

\definecolor{oursrow}{RGB}{232,242,252}
\definecolor{avgcol}{RGB}{236,236,236}

\hypersetup{pdfsubject={Preprint}}

\mlsystitlerunning{Dynamic Mixed-Precision Routing for Efficient Multi-step LLM Interaction}

\begin{document}

\twocolumn[
\mlsystitle{Dynamic Mixed-Precision Routing for Efficient\texorpdfstring{\\}{ }Multi-step LLM Interaction}

\begin{mlsysauthorlist}
\mlsysauthor{Yuanzhe Li}{ua}
\mlsysauthor{Jianing Deng}{pitt}
\mlsysauthor{Jingtong Hu}{pitt}
\mlsysauthor{Tianlong Chen}{unc}
\mlsysauthor{Song Wang}{ucf}
\mlsysauthor{Huanrui Yang}{ua}
\end{mlsysauthorlist}

\mlsysaffiliation{ua}{University of Arizona, Tucson, AZ, USA}
\mlsysaffiliation{pitt}{University of Pittsburgh, Pittsburgh, PA, USA}
\mlsysaffiliation{unc}{University of North Carolina at Chapel Hill, Chapel Hill, NC, USA}
\mlsysaffiliation{ucf}{University of Central Florida, Orlando, FL, USA}

\mlsyskeywords{LLM Agents, Quantization, Mixed Precision, Routing, Efficient Inference}

\vskip 0.3in

\begin{abstract}
Large language models (LLMs) achieve strong performance in long-horizon agentic tasks through multi-step interaction and reasoning at test time.
However, repeatedly invoking these models incurs substantial inference cost over an episode.
Post-training quantization (PTQ) reduces this cost, but using a fixed low precision throughout an episode can sharply lower task success.
We observe that interaction steps differ sharply in their sensitivity to quantization: given the same context, decoding a step at low precision turns a success into a failure at only about 1\% of the steps, which we call critical steps.
Based on this observation, we propose \textbf{Dynamic Mixed-Precision Routing (DMR)}, a framework that adaptively selects between high-precision and low-precision versions of the same LLM at each decision step.
Leveraging an entire-space factorization to jointly predict a step's divergence and its effect on the outcome, \sys{} learns to recognize these critical steps by training only a lightweight router on the model's hidden state. By reusing a shared high-precision KV cache and the hidden states, \sys{} avoids redundant context prefill and a separate text encoder, enabling precision switching with less than 1\,ms of overhead per step and adding less than 0.11\% to step latency for routing. This mixed-precision design also enables nested quantization, reducing the memory footprint by 39\% on average compared with FP16.
Across three long-horizon agentic benchmarks with nested quantization, \sys{} achieves an average $2.44\times$ decode speedup over FP16 at comparable task success rates, without fine-tuning the underlying model. On ALFWorld with GPTQ, \sys{} achieves a $1.71\times$ decode speedup with 86.9\% success, compared with 87.2\% for FP16.
\end{abstract} 

]

\printAffiliationsAndNotice{}

\section{Introduction}
\label{sec:intro}

\begin{figure*}[t]
\centering
\includegraphics[width=\textwidth]{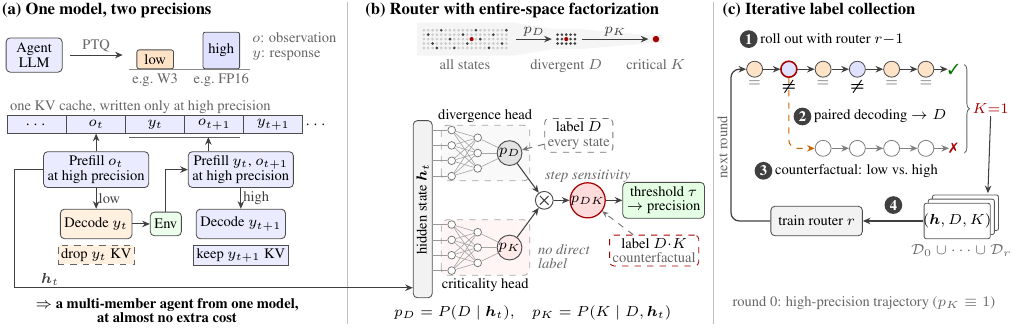}
\caption{Overview of \sys{}.
(a) Precision tiers of the same LLM share a high-precision KV cache, enabling step-level routing without redundant context prefill. Low-precision responses are re-prefilled at high precision in subsequent steps (\S\ref{sec:design}).
(b) The router predicts step criticality through an entire-space factorization of action divergence and conditional criticality, using hidden-state features and counterfactual supervision (\S\ref{sec:sup-esmm}).
(c) Iterative label collection combines agent rollouts, paired decoding, and counterfactual branching to train the router across rounds (\S\ref{sec:sup-states}).}
\label{fig:framework}
\end{figure*}

Large language models have demonstrated strong performance
in long-horizon agentic tasks involving multi-step decision
making, tool use, and environmental interaction~\cite{react,
schick2023toolformer,DBLP:journals/tmlr/WangX0MXZFA24,
shinn2023reflexion,DBLP:conf/nips/YangJWLYNP24}.
Running these agents on personal computers, workstations,
and edge devices keeps user data local and avoids remote
inference charges~\cite{powerinfer}.
Recent work develops on-device agents for phones and
desktops~\cite{ferretuilite,openphone}, alongside inference
engines tailored to their multi-step workloads~\cite{agentx,
agentxpu,edgeagent}.
For interactive local use, the key trade-off is between
task success and latency.
This latency accumulates across interaction steps as the
agent repeatedly calls the model with a growing task
context~\cite{agentx,agentxpu}.

Low-bit quantization offers a practical way to reduce the
memory footprint and inference latency of locally deployed
language models.
At small batch sizes, decoding is often memory-bound,
making weight precision an important determinant of latency.
Post-training quantization (PTQ) reduces the precision of
an existing model without fine-tuning its weights~\cite{gptq,awq},
preserves performance on many static benchmarks~\cite{llmint8,awq,jin-etal-2024-comprehensive},
and is already used in on-device language models~\cite{apple-afm}.

These efficiency gains, however, can come at a substantial
cost to agent task success~\cite{quantagent}.
On ALFWorld and ScienceWorld, fixed 3-bit execution reduces success
rates by 58 and 41 percentage points, respectively.
Using a fixed precision throughout an episode therefore
limits how much of this efficiency gain can be realized
without sacrificing task success.

\begin{figure}[t]
  \centering
  \includegraphics[width=\columnwidth]{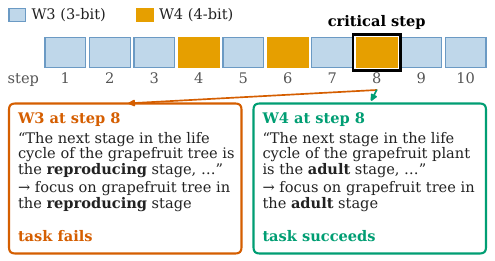}
  \caption{One ScienceWorld episode (life-cycle task, a training task of the router). Each cell is one step, colored by the precision that decoded it. At step 8, the W3 output skips the adult stage and the task fails at once, while the W4 output names it and the task succeeds two steps later. Only a few steps need W4.}
  \label{fig:Reasoning_Trajectory}
\end{figure}

\begin{figure}[t]
\centering
\includegraphics[width=\columnwidth]{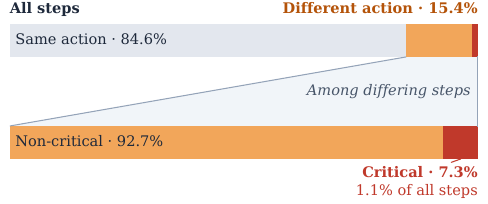}
\caption{Critical steps are rare: GPTQ W3 vs.\ FP16 on 540 ALFWorld training games.}
\label{fig:sensitivity}
\end{figure}

We observe that an agent episode is a sequence of decisions of unequal difficulty, and the step offers a natural dimension along which to allocate computation.
Examining the failed trajectories of quantized models, we find that the quantized model handles most steps well but meets a few \textit{critical steps} that are beyond its capability and lead to a wrong outcome, as illustrated in Figure~\ref{fig:Reasoning_Trajectory}.
We empirically find that critical steps are rare on ALFWorld. Given the same context, the 3-bit model selects a different action from the 16-bit model at only 15\% of the steps. Among these divergent steps, only 7\% turn an otherwise successful episode into a failure. Consequently, only about 1\% of all steps are critical (Figure~\ref{fig:sensitivity}).
This raises the question of how to allocate precision across
agent steps to preserve task success while reducing inference cost.

Building on this observation, we propose \textbf{Dynamic Mixed-Precision Routing (DMR)}, which adaptively selects among different precision tiers of the same post-training-quantized model at each agent step.
A lightweight learned router guides precision selection, allowing the agent to adjust its decoding precision throughout an episode.

\sys{} reuses a prefill at the highest precision of the set (the top tier)~\cite{pmpd} at each agent step across all
precision variants: its KV cache supports decoding at the
selected precision, and its hidden state supplies the router
input. This reuse enables low-cost precision switching
without re-prefilling the growing history or running a
separate encoder for routing~\cite{mtrouter}
(Figure~\ref{fig:framework}a,
\S\ref{sec:eval-system}).
Table~\ref{tab:positioning} highlights the distinction between \sys{} and prior work: \sys{} combines agent-step-level precision routing within a single model with cross-precision KV-cache sharing.

To train the router, \sys{} exploits how critical steps arise: a step can be critical only if the precisions output different actions, which is cheap to observe at every step, whereas criticality itself is rare and observed only through a counterfactual branch. As with clicks and conversions in recommender systems, \sys{} learns both with an entire-space factorization~\cite{esmm}.

We evaluate \sys{} on ALFWorld~\cite{alfworld}, ScienceWorld~\cite{scienceworld} and TextWorld~\cite{textworld} with nested quantization from Any-Precision LLM~\cite{anyprecision} and independent quantization from GPTQ~\cite{gptq}.
Without fine-tuning the model, \sys{} achieves comparable success rate across all benchmarks while decoding $1.71$--$3.25\times$ faster than FP16.
We make the following contributions:
\begin{itemize}[leftmargin=1.2em, itemsep=2pt]
    \item \textbf{Critical steps are rare but decisive.} We identify critical steps, the few steps at which low precision turns a successful episode into a failure, and show that they decide the success of quantized agents (Figure~\ref{fig:sensitivity}).
    \item \textbf{A lightweight router learns to find critical steps.} \sys{} trains a linear probe on the model's own hidden state
to guide precision selection at each agent step.
An entire-space factorization combines two
supervision signals: inexpensive step-level divergence and
sparse criticality labels obtained from counterfactual branches
(\S\ref{sec:supervision}).
    \item \textbf{Low-overhead precision routing through prefill reuse
and nested quantization.}
\sys{} reuses the prefill KV cache across precision tiers,
enabling low-cost tier switching, and uses the prefill
hidden state as router input without an additional embedding model.
Nested quantization stores all precision tiers together in
less memory than a single FP16 model
(\S\ref{sec:design}, \S\ref{sec:eval-system}).
\end{itemize}

\section{Background}
\label{sec:background}
\label{sec:bg-step}
\label{sec:bg-quant}

An agent episode is a sequence of steps.
At each step the agent builds a prompt from the task instruction, the interaction history and the newest observation, generates a response (a thought, an action, or both), and the environment executes the action.
Each prompt is the previous prompt followed by the previous response and the new observation, so prompts grow over an episode, and a model that keeps its KV cache across steps only prefills these added tokens; a different model switched in at a step must first prefill the history it has not cached.
On the benchmarks we evaluate, a prompt averages about 2k tokens and a response 11--35 tokens (Table~\ref{tab:workload} in Appendix~\ref{app:protocols}).
In local deployment, where agents run at small batch sizes, decoding is bound by memory bandwidth and takes 81--95\% of the model time when only the added tokens are prefilled.

Most of this memory traffic comes from the weights rather than the activations~\cite{awq}, so we apply weight-only quantization.
It stores the weights in $b$ bits and keeps the activations and the KV cache in 16-bit.
Prefill multiplies the weights with all new tokens at once (GEMM) and gains little speed, whereas decoding, a matrix--vector product per token (GEMV), speeds up as the weights shrink (Appendix~\ref{app:latency}).
We obtain the precisions by post-training quantization (PTQ), which needs no retraining and applies to any released agent checkpoint.
Under independent PTQ, each precision requires a separate checkpoint.
To reduce memory overhead, nested quantization stores a single set of weights that supports multiple precisions at runtime, with a memory footprint determined by the highest precision.
We adopt Any-Precision LLM~\cite{anyprecision} for this purpose.

\section{\sys{} Design}
\label{sec:design}

\sys{} executes each model call of an agent as shown in Figure~\ref{fig:framework}a and Algorithm~\ref{alg:step}.
This section describes the precision tiers (\S\ref{sec:design-tiers}), the per-step execution (\S\ref{sec:design-step}), the cache policy across steps (\S\ref{sec:design-kv}) and the router (\S\ref{sec:design-router}).
Section~\ref{sec:supervision} describes how the router is trained.

\para{Problem.}
We construct two precision tiers $\tiers = \{\blow, \btop\}$, $\blow < \btop$, from a single model using PTQ.
At step $t$, the agent's prompt is $x_t = x_{t-1} \oplus y_{t-1} \oplus o_t$, consisting of the previous prompt, response, and new observation. Based on the current state, our router selects a decoding tier $b_t \in \tiers$ for generating the response.
The goal is to maximize task success under a budget on expected task latency,
\begin{equation}
\max_{\pi}\ \Pr[\text{success}] \quad \text{s.t.}\quad \mathbb{E}\Big[\sum_{t=1}^{T} \mathrm{cost}(x_t, y_t, b_t)\Big] \le C,
\label{eq:objective}
\end{equation}
where the cost of a step is its model time: the top-tier prefill of the new tokens, the routing, and the decoding of the response $y_t$ at tier $b_t$.
A single threshold on the router output traces the trade-off between success and latency.

\subsection{Precision Tiers}
\label{sec:design-tiers}

\sys{} supports precision tiers constructed using either independent or nested post-training quantization (PTQ).

For nested quantization, we switch precision tiers at each decoding step by selecting the corresponding bit-planes and centroid lookup table. 
For independent PTQ, we keep the quantized linear layers of all precision tiers resident in GPU memory. These tiers share the same FP16 embedding, normalization, and output layers. 

All precision tiers share the same architecture, tokenizer, and KV-cache layout, enabling KV-cache reuse across tiers.

\subsection{High-Precision Prefill and Adaptive Decoding}
\label{sec:design-step}

At each step, \sys{} prefills the new prompt tokens at the top tier $\btop$, updating the context KV cache and computing the logits $z$ at the last prompt position. We also extract the final-norm hidden state $\hid_t$ of the last prompt token as the router input. The router then selects the decoding tier $b_t$ based on $\hid_t$ (\S\ref{sec:design-router}), which \sys{} uses to generate the remaining response tokens.
Prefill latency varies little across precision tiers, whereas lower bit widths reduce decoding latency (Appendix~\ref{app:latency}). We therefore perform prefill at the top tier and apply precision routing only during decoding.

Our high-precision prefill design provides three benefits.

\noindent\textbf{First,} switching decoding precision requires no additional context prefill because all tiers reuse the existing KV cache.

\noindent\textbf{Second,} we reuse the hidden state computed during prefill as the router input, introducing only the probe's computational overhead.

\noindent\textbf{Third,} low-precision decoding benefits from a high-precision representation of the context. With W3 decoding after a top-tier prefill (Top$\to$W3 in Table~\ref{tab:main}), the success rate increases from 29.6\% to 79.2\% on ALFWorld and from 30.3\% to 60.7\% on ScienceWorld, consistent with prior findings on high-precision prefill~\cite{pmpd}.

\begin{algorithm}[t]
\caption{One \sys{} step.}
\label{alg:step}
\begin{algorithmic}[1]
\REQUIRE context cache $\mathcal{C}$ (written at the top tier $\btop$); previous tier $b_{t-1}$ and response $y_{t-1}$; new observation $o_t$; probe $f$; threshold $\tau$
\STATE \textbf{if} $b_{t-1} = \blow$: $u \gets y_{t-1} \oplus o_t$ \hfill\COMMENT{re-prefill $y_{t-1}$}
\STATE \textbf{else}: $u \gets o_t$ \hfill\COMMENT{$y_{t-1}$ already in $\mathcal{C}$}
\STATE $(\mathcal{C}, z, \hid) \gets \textsc{Prefill}_{\btop}(u; \mathcal{C})$ \hfill\COMMENT{$\mathcal{C}$ now holds $x_t$}
\item[] \hfill\COMMENT{$z$, $\hid$: logits, final-norm state}
\STATE $s \gets 1 - f(\hid)$ \hfill\COMMENT{sufficiency score}
\STATE $b_t \gets \blow$ if $s \ge \tau$, else $\btop$
\STATE $y_t \gets \textsc{Decode}_{b_t}(\mathcal{C}, \text{first token} \sim z)$
\STATE \textbf{if} $b_t = \blow$: drop the entries of $y_t$ from $\mathcal{C}$
\STATE \textbf{return} $y_t$
\end{algorithmic}
\end{algorithm}

\subsection{High-Precision KV-Cache Management}
\label{sec:design-kv}

Agent context grows across multiple steps by appending new observations and responses, allowing the KV cache of $x_t$ to be reused in subsequent steps. However, when a response $y_t$ is generated at a low-precision tier, its cache entries are computed using low-precision weights.

To maintain a high-precision context cache, \sys{} discards these entries at the end of each step and re-prefills $y_t$ at $\btop$ together with the next observation $o_{t+1}$ (Algorithm~\ref{alg:step}, lines~1 and~3). This ensures that all context cache entries are computed at $\btop$, preventing the accumulation of low-precision cache entries across steps. It also keeps the router input $\hid_t$ consistent with the high-precision context used during probe training.

The additional overhead is small because re-prefilling only processes the short responses generated at low precision. On our test trajectories, this operation increases episode-level prefill time by only 2.1--3.8\% (\S\ref{sec:eval-system}).

\subsection{Probe-Based Precision Routing}
\label{sec:design-router}

We implement the router as a lightweight probe $f$ that takes $\hid_t$ as input. This hidden state encodes information from the task, interaction history, and current observation, thereby capturing the context needed for the upcoming decision.
Our router predicts the probability $p_{DK}$ that decoding this step at $\blow$ would be critical. We compute this probability as the product of a divergence head and a criticality head that share $\hid_t$ as input (\S\ref{sec:sup-esmm}, Figure~\ref{fig:framework}b). 
We then compute the sufficiency score $s = 1 - p_{DK} \in [0,1]$ for routing. 
We standardize the probe inputs using training-set statistics to ensure consistent feature scales.
The step decodes at $\blow$ if $s \ge \tau$ and at $\btop$ otherwise.

Since the probe reads the hidden state already computed during prefill, it needs no separate encoder or tokenization; \S\ref{sec:eval-system} compares its overhead with a text encoder (Table~\ref{tab:router-cost}).

\section{Cross-Precision Supervision}
\label{sec:supervision}

The \sys{} router estimates the criticality of each step to guide precision selection. Since critical steps are rare and labeling them requires costly counterfactual rollouts, we leverage action divergence between precision tiers as a cheaper and denser supervision signal. We integrate divergence and criticality supervision through an ESMM-inspired entire-space factorization~\cite{esmm}.

\subsection{Divergence and Criticality Labels}
\label{sec:sup-labels}

To construct both supervision labels, \sys{} generates paired responses from the same state at different precision tiers and compares the actions they produce. For a state with prompt $x$, \sys{} first prefills $x$ at $\btop$, following the deployment procedure. Each tier $b \in \tiers$ then generates a response $y_b$ using greedy decoding from the same high-precision prompt KV cache, with the first token selected from the top-tier prefill logits.
We compare free generations rather than teacher-forced likelihoods. Teacher forcing conditions the low-precision model on the top-tier response, preventing us from observing how precision-induced differences in earlier tokens affect subsequent reasoning and the final action.

We construct two supervision labels with different annotation costs: divergence and criticality. Let $a(y)$ be the action parsed from response $y$. The response of tier $b$ \emph{agrees} with the top tier if $a(y_b) = a(y_{\btop})$ and both responses terminate normally. We define divergence as $D = 1$ for disagreement and $D = 0$ otherwise.

At a divergent state, \sys{} obtains the \emph{criticality} label $K$ through counterfactual branching. We construct two branches by executing the top-tier and low-tier responses separately. Both branches then continue under the same router until the episode terminates. We define $K = 1$ when the top-tier branch succeeds and the low-tier branch fails. Non-divergent states have $K = 0$ by definition.

\subsection{Entire-Space Factorization}
\label{sec:sup-esmm}

Because criticality implies divergence, the probability that a step is critical factorizes as
\begin{equation}
\underbrace{P(D{=}1, K{=}1 \,|\, \hid_t)}_{p_{DK}} = \underbrace{P(D{=}1 \,|\, \hid_t)}_{p_D} \cdot \underbrace{P(K{=}1 \,|\, D{=}1, \hid_t)}_{p_K},
\label{eq:esmm}
\end{equation}
where $\hid_t$ is the router input of \S\ref{sec:design-router}. Directly predicting divergence or criticality presents distinct limitations. 
A divergence classifier treats every divergence as critical, unnecessarily allocating top-tier decoding to harmless divergences (at most $1.28\times$ against $1.71\times$ on ALFWorld with GPTQ, \S\ref{sec:eval-ablation}).
Directly training a criticality classifier on all states is challenging due to extreme class imbalance.
Training a conditional criticality classifier only on states with counterfactual branches mitigates the imbalance but introduces sample-selection bias: these are all divergent states, whereas the router must score every state at deployment, including the non-divergent majority that such a classifier never sees.

\sys{} models both factors over the entire space of labeled states (Figure~\ref{fig:framework}b). We use the frozen hidden state $\hid_t$ as the shared input to two prediction heads, producing $p_D$ and $p_K$, respectively. The two heads are jointly trained using divergence labels from all labeled states and joint criticality labels from states with known criticality:
\begin{equation}
\begin{aligned}
\mathcal{L} ={}& \sum_{x \in \mathcal{S}} \mathrm{BCE}\bigl(p_D(x),\; D(x)\bigr) \\
&+ \sum_{x \in \mathcal{S}_K} \mathrm{BCE}\bigl(p_D(x)\,p_K(x),\; D(x)K(x)\bigr),
\end{aligned}
\label{eq:esmm-loss}
\end{equation}
where $\mathcal{S}$ is the set of labeled states, indexed by their prompts, and $\mathcal{S}_K \subseteq \mathcal{S}$ is the set of states whose criticality is known: every non-divergent state and every divergent state with a counterfactual branch.
We supervise the criticality head indirectly through the product $p_D p_K$. This factorization also guarantees $p_{DK} \le p_D$. 
The two loss terms provide supervision at different levels of coverage. The divergence objective trains $p_D$ on all collected states, while the joint criticality objective uses counterfactual labels to supervise $p_Dp_K$. Their product thus covers all states scored at deployment, with counterfactual rollouts only at divergent states.

\subsection{Iterative State Collection}
\label{sec:sup-states}

Precision routing changes the agent's trajectory, causing the resulting state distribution to differ from that of top-tier-only execution. \sys{} therefore collects labels iteratively, in the spirit of DAgger~\cite{dagger}. In round~0, we run the agent at the top tier on training tasks and label divergence at every visited state. Since no router is available for counterfactual branching, we train an initial divergence-only router by setting $p_K \equiv 1$ in Equation~\ref{eq:esmm}. The router then selects decoding precision based on the predicted divergence probability.

In round $r$, we run the agent on training tasks using the router from round $r-1$. We label divergence at every visited state and collect counterfactual branches at divergent states. We then retrain the router on the accumulated data from all rounds. Divergent states from round~0 contribute only to the first term of Equation~\ref{eq:esmm-loss}. All experiments use a single round ($r = 1$).

To account for correlations among states from the same task, we select the regularization strength using held-out tasks and refit the router on all training states. For offline evaluation, we use task-grouped cross-validation. Divergence labels reuse the state's prefill, while counterfactual branching requires an additional rollout only at divergent states. All labels and router fitting use the training split, with evaluation performed on the disjoint test split.

\section{Implementation}
\label{sec:impl}

\sys{} is implemented on top of PyTorch and Hugging Face Transformers.
For nested quantization, \sys{} uses an adapted version of the bit-plane GEMV kernel from Any-Precision LLM to support CUDA Graph execution.
The engine exposes two calls per step: a prefill at the top tier and a decode at the selected tier.
Each benchmark's agent loop invokes the router between these calls.

\para{Cache management.}
The engine tracks the precision tier used to compute each span of the KV cache.
At each prefill, \sys{} identifies the longest prefix of the new prompt that the cache holds at $\btop$ and reuses its KV entries.
We then prefill all remaining tokens at $\btop$, including responses previously decoded at the low tier.
This prefix matching eliminates the need for a separate pass to discard low-tier entries (Algorithm~\ref{alg:step}, line~7).

\para{Switching tiers.}
For Any-Precision, each linear layer stores the bit-planes of the top tier and a centroid lookup table for each tier.
The kernel selects the first $b$ bit-planes and the corresponding lookup table when decoding at tier $b$.
For GPTQ, \sys{} replaces the seven linear layers in each Transformer block with layers that hold the FP16 weight and, for each lower tier, the GPTQ integer codes, zero points, and scales. The embedding, normalization, and output layers are shared across precision tiers.
In both cases, precision switching only updates a per-layer precision flag, without copying or allocating weights.
The engine checks at runtime that weight pointers remain unchanged across precision switches.

\para{Router at runtime.}
We capture the final-norm hidden state of the last prompt token using a read-only forward hook during top-tier prefill. 
We execute the router on the CPU. The hidden state is transferred from GPU to host memory and standardized using training-set statistics before being processed by the two linear heads.

\para{Prefill and decode paths.}
We implement prefill using PyTorch's scaled-dot-product attention. Full prefill uses FlashAttention, while incremental prefill uses the memory-efficient attention kernel with an explicit causal mask.

We capture a separate CUDA graph for each precision tier and replay the corresponding graph during decoding.
All graphs share a static FP16 KV buffer and use FlashAttention's KV-cache kernel.

For latency measurement, the decode graph fuses the QKV and gate--up projections and uses fused normalization, rotary, and activation operators from vLLM~\cite{vllm}.
For GPTQ, success rates use the exact GPTQ weights, while decoding latency is measured using vLLM's 3-bit kernel.
Appendix~\ref{app:latency} validates the modeled decoding latency against wall-clock measurements.

\section{Evaluation}
\label{sec:eval}

We evaluate \sys{} by answering four questions.
\textbf{Q1:} Does step-level precision routing improve the success--latency trade-off over fixed precisions and heuristic schedules (\S\ref{sec:eval-main}, \S\ref{sec:eval-ptq})?
\textbf{Q2:} What system-level benefits does precision routing offer compared with routing across separate models (\S\ref{sec:eval-system})?
\textbf{Q3:} Does the router identify critical steps based on the current state rather than relying on step position (\S\ref{sec:eval-learn})?
\textbf{Q4:} Which design choices matter most (\S\ref{sec:eval-ablation})?

\subsection{Experimental Setup}
\label{sec:eval-setup}

\para{Benchmarks and precision sets.}
We evaluate \sys{} on three agent benchmarks: ALFWorld~\cite{alfworld}, ScienceWorld~\cite{scienceworld}, and TextWorld cooking games~\cite{textworld}. We use Qwen3-4B~\cite{qwen3} for ALFWorld and TextWorld, and ER1-7B from Embodied Planner-R1~\cite{embodiedr1} for ScienceWorld (Appendix~\ref{app:protocols}).

For nested quantization, Any-Precision LLM~\cite{anyprecision} provides $\{\text{W3}, \text{W8}\}$ for Qwen3-4B and $\{\text{W3}, \text{W4}\}$ for ER1-7B. Both top tiers achieve success rates comparable to FP16. For independent PTQ, we use GPTQ~\cite{gptq} with $\{\text{W3}, \text{FP16}\}$ on ALFWorld.

\para{Baselines.}
\emph{Fixed precision} decodes every step at a fixed precision tier. \emph{Random} selects W3 at each step with a probability calibrated to match \sys{}'s W3 token share (averaged over three random seeds). The \emph{step schedule} uses the top tier for the first $k \in \{2, 4, 6\}$ steps and W3 for all subsequent steps.

\para{Metrics.}
We report success rate, W3 token share, and decode time per task. We estimate decode time using the measured time per output token (TPOT) of each precision tier and the number of tokens decoded at that tier. TPOT is measured on an RTX A6000 with batch size one and a 2k-token context. We validate the estimated decode time against wall-clock measurements, which match the estimated speedups within $0.04\times$ (Appendix~\ref{app:latency}), and measure prefill and routing overhead separately in \S\ref{sec:eval-system}. Since decoding precision can affect episode length, we evaluate latency at the task level rather than the step level.

\subsection{Main Results}
\label{sec:eval-main}

Table~\ref{tab:main} compares all methods in the nested setting. Across all benchmarks, \sys{} decodes 65--77\% of tokens at W3 while achieving success rates within 2.9 percentage points of the top tier. On average, \sys{} recovers 81\% of the success gap between Top$\to$W3 and the top tier, compared with at most 48\% for the heuristic baselines.

\begin{figure*}[t]
\centering
\includegraphics[width=\textwidth]{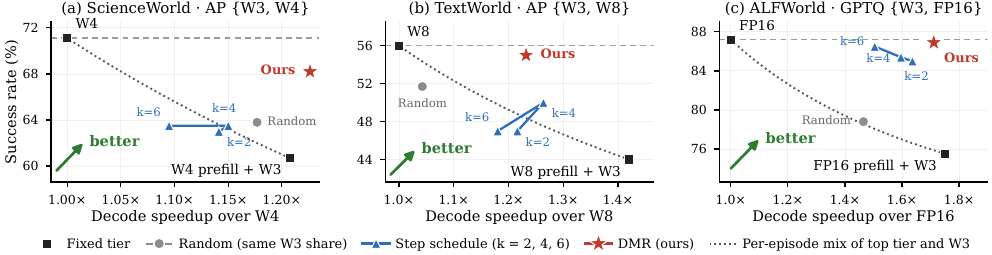}
\caption{Success rate versus episode decode speedup relative to the top precision tier. (a)~ScienceWorld with ER1-7B and Any-Precision $\{\text{W3}, \text{W4}\}$; (b)~TextWorld cooking with Qwen3-4B and Any-Precision $\{\text{W3}, \text{W8}\}$; (c)~ALFWorld with Qwen3-4B and GPTQ $\{\text{W3}, \text{FP16}\}$. \sys{} lies on the success--speedup frontier in all three settings. Dashed lines indicate top-tier success rates, dotted lines indicate per-episode mixtures of the top tier and W3, and squares mark the top-tier and Top$\to$W3 baselines.}
\label{fig:pareto}
\end{figure*}

\para{ScienceWorld.}
On ScienceWorld (Figure~\ref{fig:pareto}a), decoding every step at W3 after a W4 prefill achieves a $1.21\times$ speedup over W4 but reduces success by 10.4 percentage points. \sys{} achieves 68.2\% success at a $1.23\times$ speedup, recovering approximately 70\% of this loss.

Despite decoding some steps at W4, \sys{} achieves a lower decode time per task than W4$\to$W3 (3.39 vs.\ 3.44\,s), as low-precision decoding can increase episode length. \sys{} also outperforms all heuristic baselines in success rate by 4.4--5.2 percentage points.

\para{TextWorld.}
On TextWorld (Figure~\ref{fig:pareto}b), \sys{} achieves a 55.0\% success rate, close to W8's 56.0\%, while decoding 65\% of tokens at W3 and achieving a $1.23\times$ speedup over W8. In contrast, W8$\to$W3 achieves only 44.0\% success. \sys{} outperforms all heuristic baselines by 3.3--8.0 percentage points. The step schedule with $k=4$ is the only slightly faster heuristic ($1.26\times$), but it recovers only 50\% of the success gap, compared with 92\% for \sys{}.

\para{ALFWorld.}
On ALFWorld, decoding every step at W3 after a W8 prefill achieves a $1.37\times$ speedup over W8 but reduces success by 9.1 percentage points. In contrast, \sys{} decodes 76\% of tokens at W3 while achieving 86.5\% success at a $1.40\times$ speedup, only 1.8 percentage points below W8 and 7.3 points above W8$\to$W3.

\sys{} outperforms all heuristic baselines by 1.1--4.5 percentage points. The step schedules with $k=2$ and $k=4$ are slightly faster and come closest to \sys{} in success rate, consistent with the observation that two-thirds of ALFWorld's critical steps occur within the first two steps (Figure~\ref{fig:position}). However, they recover only 64\% and 68\% of the success gap, respectively, compared with 80\% for \sys{}.

\begin{table*}[t]
\centering
\caption{Main results with nested quantization. We use ER1-7B with $\{\text{W3}, \text{W4}\}$ on ScienceWorld and Qwen3-4B with $\{\text{W3}, \text{W8}\}$ on TextWorld and ALFWorld. Succ. denotes task success rate (\%), Speedup is episode decode speedup relative to FP16 averaged over test tasks, W3 denotes the share of tokens decoded at W3, and Recov. measures the fraction of the success gap between Top$\to$W3 and the top tier recovered by each method. Fixed-precision baselines prefill at their own tier, while all other methods use top-tier prefill. Random results are averaged over three seeds. Bold and underlined values indicate the best and second-best routing methods, respectively.}

\label{tab:main}
\footnotesize
\setlength{\tabcolsep}{4.2pt}
\begin{tabular}{llccccccccc>{\columncolor{avgcol}}c>{\columncolor{avgcol}}c}
\toprule
 & & \multicolumn{3}{c}{ScienceWorld} & \multicolumn{3}{c}{TextWorld} & \multicolumn{3}{c}{ALFWorld} & \multicolumn{2}{>{\columncolor{avgcol}}c}{\textbf{Average}} \\
\cmidrule(lr){3-5}\cmidrule(lr){6-8}\cmidrule(lr){9-11}\cmidrule(lr){12-13}
 & Method & Succ. & Speedup & W3 & Succ. & Speedup & W3 & Succ. & Speedup & W3 & Recov. & Speedup \\
\midrule
\multirow{3}{*}{Fixed} & FP16 & 71.6 & 1.00$\times$ & -- & 55.0 & 1.00$\times$ & -- & 87.2 & 1.00$\times$ & -- & -- & 1.00$\times$ \\
 & Top tier Precision & 71.1 & 2.65$\times$ & 0\% & 56.0 & 1.62$\times$ & 0\% & 88.3 & 1.49$\times$ & 0\% & 100\% & 1.92$\times$ \\
 & W3 & 30.3 & 1.54$\times$ & 100\% & 2.0 & 2.48$\times$ & 100\% & 29.6 & 1.24$\times$ & 100\% & -- & 1.75$\times$ \\
\midrule
Top prefill & Top $\to$ W3 & 60.7 & 3.20$\times$ & 100\% & 44.0 & 2.29$\times$ & 100\% & 79.2 & 2.05$\times$ & 100\% & 0\% & 2.51$\times$ \\
\midrule
\multirow{6}{*}{Routing} & Random & \underline{63.8} & \underline{3.12$\times$} & 77\% & \underline{51.7} & 1.68$\times$ & 66\% & 82.0 & 1.82$\times$ & 75\% & 42\% & 2.21$\times$ \\
 & \textit{Step schedule} &  &  &  &  &  &  &  &  &  &  &  \\
 & \quad $k=2$ & 63.0 & 3.02$\times$ & 84\% & 47.0 & 1.97$\times$ & 93\% & 85.0 & \textbf{2.18$\times$} & 73\% & 37\% & 2.39$\times$ \\
 & \quad $k=4$ & 63.5 & 3.05$\times$ & 69\% & 50.0 & \textbf{2.04$\times$} & 84\% & \underline{85.4} & \underline{2.14$\times$} & 68\% & \underline{48\%} & \underline{2.41$\times$} \\
 & \quad $k=6$ & 63.5 & 2.90$\times$ & 57\% & 47.0 & 1.91$\times$ & 79\% & 84.3 & 2.01$\times$ & 64\% & 36\% & 2.27$\times$ \\
\rowcolor{oursrow} & \sys{} & \textbf{68.2} & \textbf{3.25$\times$} & 77\% & \textbf{55.0} & \underline{1.99$\times$} & 65\% & \textbf{86.5} & 2.09$\times$ & 76\% & \textbf{81\%} & \textbf{2.44$\times$} \\
\bottomrule
\end{tabular}
\end{table*}

\subsection{Independent PTQ}
\label{sec:eval-ptq}

We extend the evaluation to independent GPTQ quantization on ALFWorld, using W3 and FP16 precision tiers (Table~\ref{tab:gptq}, Figure~\ref{fig:pareto}c). Decoding every step at W3 after an FP16 prefill achieves a $1.75\times$ speedup over FP16 but reduces success by 11.7 percentage points.

\sys{} achieves 86.9\% success, only one game below FP16 (87.2\%) across 274 test games, while delivering a $1.71\times$ decode speedup, close to the $1.75\times$ achieved by FP16$\to$W3.

Compared with the fastest step schedule ($k=2$), \sys{} reduces decode time by 0.13\,s per task and succeeds on five additional games (86.9\% vs.\ 85.0\%). Random routing with the same W3 token share achieves only 78.8\% success, 8.0 percentage points below \sys{}, while requiring 0.48\,s more decode time per task.

\begin{table}[t]
\centering
\caption{Independent PTQ on ALFWorld (GPTQ, $\{\text{W3}, \text{FP16}\}$, FP16 prefill for routing methods). Columns as in Table~\ref{tab:main}.}
\label{tab:gptq}
\footnotesize
\setlength{\tabcolsep}{4pt}
\begin{tabular}{lccc}
\toprule
Method & Succ. & Speedup & W3 \\
\midrule
FP16                 & 87.2 & 1.00$\times$ & 0\% \\
W3                   & 52.6 & 1.75$\times$ & 100\% \\
FP16 $\to$ W3        & 75.5 & 1.75$\times$ & 100\% \\
\midrule
Random               & 78.8 & 1.46$\times$ & 79\% \\
\textit{Step schedule} & & & \\
\quad $k=2$          & 85.0 & \underline{1.64$\times$} & 75\% \\
\quad $k=4$          & 85.4 & 1.60$\times$ & 70\% \\
\quad $k=6$          & \underline{86.5} & 1.50$\times$ & 65\% \\
\rowcolor{oursrow} \sys{} & \textbf{86.9} & \textbf{1.71$\times$} & 78\% \\
\bottomrule
\end{tabular}
\end{table}

\subsection{Precision Routing versus Model Routing}
\label{sec:eval-system}

We compare precision routing within a single model against routing across separate models along three dimensions: precision switching overhead, GPU memory footprint, and router inference overhead (Figure~\ref{fig:separate} in Appendix~\ref{app:switch-traj}).

\para{Precision switching overhead.}
All precision tiers in \sys{} share the KV cache computed during top-tier prefill, allowing precision switching without recomputing the context. On an RTX A6000, switching tiers incurs only 0.26--0.40\,ms of host-side overhead, with less than 0.04\,ms of additional GPU time. This switching overhead does not increase with context length in our measurements (Table~\ref{tab:router-cost}).

\begin{figure}[t]
\centering
\includegraphics[width=\columnwidth]{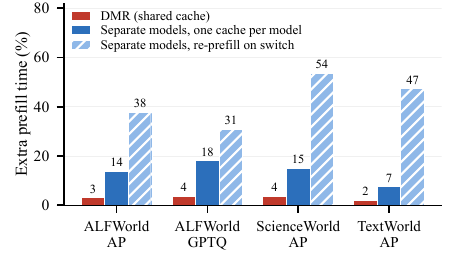}
\caption{Additional episode-level prefill time relative to top-tier-only execution. We compare \sys{} with a shared KV cache against separate-model routing with either one cache per model or context re-prefilling upon switching. All configurations follow \sys{}'s routing decisions on the same test trajectories (Appendix~\ref{app:switch-traj}).}
\label{fig:switch-traj}
\end{figure}

We further evaluate episode-level overhead by replaying \sys{}'s routing decisions on its test trajectories (Appendix~\ref{app:switch-traj}). Separate-model routing increases prefill time by 7.5--18\% when each model maintains its own cache, and by 31--54\% when the selected model re-prefills the context upon switching. In contrast, \sys{} adds only 2.1--3.8\% by re-prefilling low-precision responses at the top tier (Figure~\ref{fig:switch-traj}).

\para{GPU memory footprint.}
By sharing bit-planes and a single KV cache across precision tiers, nested \sys{} requires less GPU memory than FP16 alone and approximately one-third to one-half of separate FP16 and W3 deployments (Table~\ref{tab:memory}). Independent PTQ retains an additional set of quantized weights but still uses less memory than separate-model deployment.

\begin{table}[t]
\centering
\caption{Peak GPU memory usage (GB) across deployment configurations at batch size one, including the CUDA context and full-context KV caches (8292 tokens for 4B and 4296 for 7B). $^\dagger$Computed from the measured FP16 deployment: plus the 3-bit weights (independent), or plus a second W3 deployment with its own KV cache (separate).}
\label{tab:memory}
\footnotesize
\setlength{\tabcolsep}{4pt}
\begin{tabular}{lcc}
\toprule
Deployment & 4B & 7B \\
\midrule
FP16 only                                          & 11.4 & 16.0 \\
Separate FP16 and W3 models$^\dagger$              & 17.3 & 22.3 \\
\sys{}, independent $\{\text{W3}, \text{FP16}\}$$^\dagger$ & 12.7 & 18.4 \\
\sys{}, nested $\{\text{W3}, \text{W4}\}$          & 6.4 & 7.1 \\
\sys{}, nested $\{\text{W3}, \text{W8}\}$          & 8.8 & -- \\
\bottomrule
\end{tabular}
\end{table}

\para{Router inference overhead.}
\sys{} reuses the final-norm hidden state computed during top-tier prefill for routing, avoiding an additional encoder or context processing. The two linear heads contain 5.1K parameters for Qwen3-4B and 7.2K for ER1-7B. Including the GPU-to-CPU transfer, routing takes only 0.19--0.31\,ms per step, compared with 48--61\,ms for the Qwen3-Embedding-0.6B text encoder used in MTRouter~\cite{mtrouter} on the same GPU and agent histories (Table~\ref{tab:router-cost}).

\begin{table}[t]
\centering
\caption{Per-step overhead on an RTX A6000. Re-prefill latency is measured at the agents' average context length of 2k tokens. Encoder latency uses Qwen3-Embedding-0.6B, as in MTRouter~\cite{mtrouter}, averaged over 1{,}000 steps per benchmark from the test trajectories. Probe latency includes GPU-to-CPU hidden-state transfer.}

\label{tab:router-cost}
\footnotesize
\setlength{\tabcolsep}{4pt}
\begin{tabular}{lcc}
\toprule
Per step & Separate models & \sys{} \\
\midrule
Switching & re-prefill, 218--324\,ms & 0.26--0.40\,ms \\
Router input & encoder, 48--61\,ms & probe, 0.19--0.31\,ms \\
\bottomrule
\end{tabular}
\end{table}

\subsection{What Does the Router Learn?}
\label{sec:eval-learn}

We investigate whether \sys{} learns to identify critical steps from the current state or merely relies on step position. In long-horizon tasks, critical steps can occur at different positions depending on the task and execution trajectory. We therefore compare our router against position-based scheduling, inspired by PMPD's precision scheduling within a generation~\cite{pmpd}.

We evaluate both approaches on labeled training states using out-of-fold router predictions from task-grouped cross-validation (\S\ref{sec:supervision}).

\begin{figure}[t]
\centering
\includegraphics[width=\columnwidth]{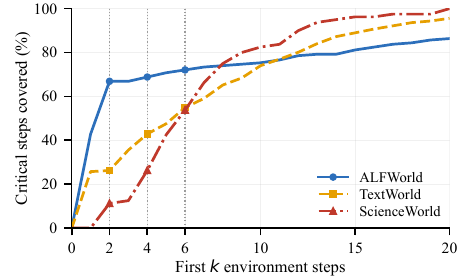}
\caption{Cumulative distribution of critical training states across environment steps for ALFWorld, TextWorld, and ScienceWorld. The curves indicate the fraction of critical states covered by allocating top-tier decoding to the first $k$ steps. Dotted lines mark $k=2,4,6$.}
\label{fig:position}
\end{figure}

\para{Critical-step positions vary across tasks.}
The distribution of critical steps varies across tasks (Figure~\ref{fig:position}). On ALFWorld, two-thirds of critical steps occur within the first two steps, whereas 56\% occur at steps 4--9 on ScienceWorld and 75\% occur after the first two steps on TextWorld. By allocating top-tier decoding to the first $k$ steps, step scheduling approaches \sys{}'s success rate on ALFWorld (85.4\% vs.\ 86.5\%) but falls behind by 4.7 and 5.0 percentage points on ScienceWorld and TextWorld, respectively (Table~\ref{tab:main}).

\begin{figure}[t]
\centering
\includegraphics[width=\columnwidth]{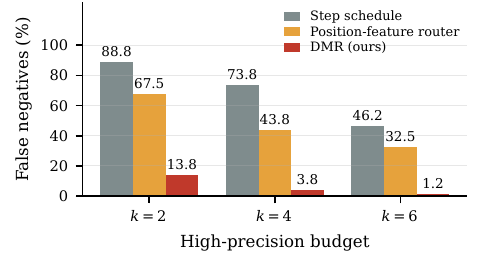}
\caption{False negatives on ScienceWorld under matched high-precision budgets. We report the FN rate, i.e., the fraction of 80 critical training steps decoded at W3 (lower is better), by step schedules, the position-feature router, and \sys{}. Each budget matches the W4 step share of the corresponding step schedule ($k=2,4,6$), while both routers independently select which steps use W4.}
\label{fig:leak}
\end{figure}

On ScienceWorld, the $k=2$ step schedule allocates most W4 calls to states where the two precision tiers agree, leaving 89\% of critical states decoded at W3, i.e., as false negatives (FN). At the same W4 usage rate, \sys{} has an FN rate of only 13.8\% (Figure~\ref{fig:leak}). The advantage persists at the W4 usage rates of the $k=4$ and $k=6$ schedules: \sys{} has FN rates of only 3.8\% and 1.2\%, compared with 74\% and 46\% for the corresponding schedules.

\para{Position features cannot find the critical steps.}
We evaluate a position-feature router by replacing the hidden-state input with the environment step index, indicators for the first two steps, and prompt length. We retain the same two-head architecture, supervision labels, and cross-validation protocol as \sys{}.

The position-feature router achieves an AUC of 0.530 in distinguishing critical from harmless divergences, compared with 0.745 for \sys{}. At the same W3 call share, the position-feature router decodes 51.7 of the 80 critical states at W3, compared with only 8.7 for \sys{} (averaged over three seeds).

\para{Budget stability across task distributions.}
We evaluate whether the routing threshold maintains a consistent low-precision usage rate across training and test tasks. We calibrate the thresholds of both \sys{} and the position-feature router on training states, then compare their W3 call shares on training and test trajectories. \sys{} maintains a similar W3 call share (80.5\% on training states vs.\ 80.1\% on test tasks), whereas the position-feature router shifts from 70.6\% to 84.8\%. This larger shift is associated with longer test episodes containing more late-stage steps, highlighting the greater budget stability of hidden-state routing.

\subsection{Ablations}
\label{sec:eval-ablation}

Criticality supervision improves routing efficiency. On ALFWorld with GPTQ, a divergence-only router allocates FP16 decoding to harmless divergences, achieving at most $1.28\times$ speedup versus $1.71\times$ for \sys{} at a higher success rate (86.1\% vs.\ 86.9\%; Table~\ref{tab:round0} in Appendix~\ref{app:ablations}).

Table~\ref{tab:ablation-factor} compares router factorizations trained on the same states and folds: a mixed-label head trained on round-0 divergence and round-1 criticality labels, a criticality-only head, separately trained divergence and conditional criticality heads, and our jointly trained entire-space factorization.
The mixed-label head performs near chance (AUC 0.448--0.535), whereas the other structures achieve substantially higher AUC.
At a fixed W3 call share, the entire-space factorization achieves the lowest or near-lowest FN across all three settings. On ScienceWorld, the criticality-only head attains a higher AUC but routes more critical states to W3 than \sys{} (11.3 vs.\ 8.7).

\begin{table}[t]
\centering
\caption{Offline comparison of router factorizations. AUC measures discrimination between critical and harmless divergences among states with counterfactual branches. FN counts critical states routed to W3 at a fixed W3 call share. Values are means over three cross-validation splits. Bold and underlined values indicate the best and second-best results in each column.}
\label{tab:ablation-factor}
\small
\setlength{\tabcolsep}{3pt}
\begin{tabular}{lcccccc}
\toprule
 & \multicolumn{2}{c}{ALF GPTQ} & \multicolumn{2}{c}{ALF AP} & \multicolumn{2}{c}{SciWorld AP} \\
\cmidrule(lr){2-3}\cmidrule(lr){4-5}\cmidrule(lr){6-7}
Structure & AUC & FN & AUC & FN & AUC & FN \\
\midrule
Mixed label            & 0.511 & 20.7 & 0.535 & 28.3 & 0.448 & 27.3 \\
Criticality only       & \underline{0.731} & 10.7 & \underline{0.623} & 20.3 & \textbf{0.779} & 11.3 \\
Separate heads         & 0.730 & \underline{7.7}  & 0.612 & \textbf{11.0} & 0.709 & \underline{10.0} \\
\rowcolor{oursrow} Entire space (\sys{})  & \textbf{0.738} & \textbf{7.0}  & \textbf{0.627} & \underline{11.3} & \underline{0.745} & \textbf{8.7} \\
\bottomrule
\end{tabular}
\end{table}

\section{Related Work}
\label{sec:related}

Table~\ref{tab:positioning} compares \sys{} with prior work on runtime precision adaptation and model routing. Existing precision adaptation methods operate at the request level~\cite{anyprecision,anybcq,matquant,morphserve} or within a generation~\cite{pmpd,dpllm,qspec}. In contrast, fine-grained routing at the reasoning-step or agent-turn level has primarily focused on selecting among separate models~\cite{steer,trim,routerr1,mtrouter,hera}.

\begin{table}[t]
\centering
\caption{Methods that choose a precision or a model for LLM inference. \emph{One model}: the candidates are precisions of a single model. \emph{Shared KV}: a newly chosen candidate decodes from the existing cache without re-prefilling (--: no switch within a request). \emph{Decided by}: what the choice depends on.}
\label{tab:positioning}
\footnotesize
\setlength{\tabcolsep}{3pt}
\resizebox{\columnwidth}{!}{%
\begin{tabular}{llccl}
\toprule
Method & \makecell[l]{Choice\\made per} & \makecell{One\\model} & \makecell{Shared\\KV} & Decided by \\
\midrule
\makecell[l]{Any-Precision,\\AnyBCQ, MatQuant} & request & \checkmark & --         & requested bit width \\
MorphServe                & load period          & \checkmark & \checkmark & server load \\
PMPD                      & decode phase         & \checkmark & \checkmark & decoding progress \\
DP-LLM                    & token $\times$ layer & \checkmark & \checkmark & layer error \\
QSpec                     & draft / verify       & \checkmark & \checkmark & fixed roles \\
\addlinespace
R2R                       & token                & $\times$   & $\times$   & token divergence \\
STEER, TRIM               & reasoning step       & $\times$   & $\times$   & confidence, PRM \\
Router-R1                 & routing round        & $\times$   & $\times$   & router LLM \\
MTRouter                  & agent turn           & $\times$   & $\times$   & encoded history \\
\midrule
\rowcolor{oursrow}
\sys{} (ours)             & agent step           & \checkmark & \checkmark & step criticality \\
\bottomrule
\end{tabular}}
\end{table}

\para{Quantization and runtime precision.}
\sys{} builds on existing quantization methods, such as GPTQ~\cite{gptq} and Any-Precision~\cite{anyprecision}, focusing on runtime precision selection.
Prior work explores precision adaptation at different granularities. PMPD~\cite{pmpd} progressively reduces decoding precision within a generation, while DP-LLM~\cite{dpllm} dynamically assigns layer-wise precision and QSpec~\cite{qspec} combines low-precision drafting with higher-precision verification.
At the serving level, MorphServe~\cite{morphserve} adapts quantized layers to workload changes, while Mix-Quant~\cite{mixquant} explores low-precision prefill with high-precision decoding for agent workloads.
In contrast, \sys{} selects decoding precision at each agent step and maintains a high-precision context cache across steps.

\para{LLM routing.}
Query-level routers select models based on input difficulty, preference data, or learned query--model representations~\cite{frugalgpt,hybridllm,routerdc,routellm,embedllm}.
Fine-grained model routing has also been explored during multi-step reasoning and agent interactions. STEER~\cite{steer} uses confidence signals, while TRIM~\cite{trim} employs process rewards to route reasoning steps between smaller and larger models. Router-R1~\cite{routerr1} learns multi-round model routing and aggregation, whereas MTRouter~\cite{mtrouter} selects models at each agent turn based on embedded interaction histories.
For long-horizon agents, Hera~\cite{hera} routes each agent step between a device and a cloud model with a router first trained on whether their actions agree. Recent work further extends model routing to agent-level decisions, requests, and subtasks~\cite{redact,agentcompress,rsirouter}. At the token level, R2R~\cite{r2r} selectively routes path-divergent tokens between small and large models.

Hidden-state-based routing has also been explored for model selection~\cite{prefillrouter}, reasoning-mode selection~\cite{selfroute}, and agent deliberation~\cite{geomdiv}.
In \sys{}, the router uses hidden states from the top-tier prefill, whose KV cache is reused by the selected decoding tier, avoiding an additional encoding or prefill pass.

\section{Discussion and Limitations}
\label{sec:limitations}

\para{Operating regime.}
\sys{} targets latency-sensitive, small-batch agent inference, where decoding is memory-bandwidth-bound and low-bit weights give the largest speedups.
The gain shrinks as the batch grows but persists: at batch size eight, \sys{} with Any-Precision still decodes $1.10\times$ and $1.09\times$ faster than the top tier on ALFWorld and ScienceWorld, and $1.24\times$ and $1.66\times$ faster than FP16 (Appendix~\ref{app:latency}).
Since \sys{} retains top-tier prefill, its end-to-end speedup also depends on the fraction of time spent decoding.

\para{Memory trade-offs.}
Nested quantization keeps all tiers within the footprint of the top tier, but its top tier is itself a quantized format (W8 or W4 here); independent PTQ allows an FP16 top tier at the cost of a second, quantized copy of the weights (12.7 vs.\ 11.4\,GB for Qwen3-4B).

\para{Prompt continuity.}
Cross-step KV-cache reuse is most effective when agent prompts preserve their previous token prefix.
For protocols that rewrite or truncate the history, \sys{} can still operate by recomputing the affected context, with reduced cache-reuse benefits.

\section{Conclusion}
\label{sec:conclusion}

Our results show that an agent's precision requirements vary across steps, even within the same model.
Low- and high-precision execution diverge on relatively few steps, and only a subset of these divergences affects task success. Building on this observation, we propose \sys{}, which predicts step criticality from prefill hidden states by jointly modeling action divergence and its impact on task success.
The predicted criticality then guides decoding precision selection at each agent step. By reusing a shared high-precision KV cache, \sys{} enables low-overhead precision switching without an additional routing encoder or re-prefilling the shared context.
Across three agent benchmarks, \sys{} achieves $1.71$--$3.25\times$ faster decoding than FP16 at comparable task success rates, without fine-tuning the underlying model.

\section*{Acknowledgments}
This work used the Delta system at the National Center for Supercomputing Applications (NCSA) [award OAC 2005572] through allocation CIS260333 from the Advanced Cyberinfrastructure Coordination Ecosystem: Services \& Support (ACCESS) program~\cite{boerner2023access}, which is supported by U.S. National Science Foundation grants \#2138259, \#2138286, \#2138307, \#2137603, and \#2138296.

\bibliography{references}

\begin{thebibliography}{51}
\providecommand{\natexlab}[1]{#1}
\providecommand{\url}[1]{\texttt{#1}}
\expandafter\ifx\csname urlstyle\endcsname\relax
  \providecommand{\doi}[1]{doi: #1}\else
  \providecommand{\doi}{doi: \begingroup \urlstyle{rm}\Url}\fi

\bibitem[Boerner et~al.(2023)Boerner, Deems, Furlani, Knuth, and
  Towns]{boerner2023access}
Boerner, T.~J., Deems, S., Furlani, T.~R., Knuth, S.~L., and Towns, J.
\newblock {ACCESS}: Advancing innovation: {NSF}'s advanced cyberinfrastructure
  coordination ecosystem: Services \& support.
\newblock In \emph{Practice and Experience in Advanced Research Computing
  (PEARC)}, pp.\  173--176. ACM, 2023.
\newblock \doi{10.1145/3569951.3597559}.

\bibitem[Chen et~al.(2025)Chen, Tan, Kouris, Lee, Fan, and Venieris]{pmpd}
Chen, H.~M., Tan, F., Kouris, A., Lee, R., Fan, H., and Venieris, S.~I.
\newblock Progressive mixed-precision decoding for efficient {LLM} inference.
\newblock In \emph{International Conference on Learning Representations
  (ICLR)}, pp.\  37767--37783, 2025.
\newblock URL
  \url{https://proceedings.iclr.cc/paper_files/paper/2025/hash/5df4313ecd4875931fbdacc486cc1fcf-Abstract-Conference.html}.

\bibitem[Chen et~al.(2024{\natexlab{a}})Chen, Zaharia, and Zou]{frugalgpt}
Chen, L., Zaharia, M., and Zou, J.
\newblock {FrugalGPT}: How to use large language models while reducing cost and
  improving performance.
\newblock \emph{Transactions on Machine Learning Research (TMLR)},
  2024{\natexlab{a}}.
\newblock ISSN 2835-8856.
\newblock URL \url{https://openreview.net/forum?id=cSimKw5p6R}.

\bibitem[Chen et~al.(2024{\natexlab{b}})Chen, Jiang, Lin, Kwok, and
  Zhang]{routerdc}
Chen, S., Jiang, W., Lin, B., Kwok, J.~T., and Zhang, Y.
\newblock {RouterDC}: Query-based router by dual contrastive learning for
  assembling large language models.
\newblock In \emph{Advances in Neural Information Processing Systems
  (NeurIPS)}, volume~37, pp.\  66305--66328. Curran Associates, Inc.,
  2024{\natexlab{b}}.
\newblock \doi{10.52202/079017-2120}.

\bibitem[Chung et~al.(2026)Chung, Shin, Kim, and Rhu]{agentx}
Chung, J., Shin, B., Kim, J., and Rhu, M.
\newblock {Agent-X}: Full pipeline acceleration of on-device {AI} agents.
\newblock In \emph{Proceedings of the 24th Annual International Conference on
  Mobile Systems, Applications and Services (MobiSys)}, pp.\  144--157. ACM,
  2026.
\newblock \doi{10.1145/3745756.3809195}.

\bibitem[C{\^o}t{\'e} et~al.(2019)C{\^o}t{\'e}, K{\'a}d{\'a}r, Yuan, Kybartas,
  Barnes, Fine, Moore, Hausknecht, El~Asri, Adada, Tay, and
  Trischler]{textworld}
C{\^o}t{\'e}, M.-A., K{\'a}d{\'a}r, {\'A}., Yuan, X., Kybartas, B., Barnes, T.,
  Fine, E., Moore, J., Hausknecht, M., El~Asri, L., Adada, M., Tay, W., and
  Trischler, A.
\newblock {TextWorld}: A learning environment for text-based games.
\newblock In \emph{Computer Games (Communications in Computer and Information
  Science)}, pp.\  41--75. Springer, 2019.
\newblock \doi{10.1007/978-3-030-24337-1_3}.

\bibitem[Dettmers et~al.(2022)Dettmers, Lewis, Belkada, and
  Zettlemoyer]{llmint8}
Dettmers, T., Lewis, M., Belkada, Y., and Zettlemoyer, L.
\newblock {GPT3.int8()}: 8-bit matrix multiplication for transformers at scale.
\newblock In \emph{Advances in Neural Information Processing Systems
  (NeurIPS)}, volume~35, pp.\  30318--30332. Curran Associates, Inc., 2022.
\newblock \doi{10.52202/068431-2198}.
\newblock Also known as {LLM.int8()}, arXiv:2208.07339.

\bibitem[Ding et~al.(2024)Ding, Mallick, Wang, Sim, Mukherjee, R{\"u}hle,
  Lakshmanan, and Awadallah]{hybridllm}
Ding, D., Mallick, A., Wang, C., Sim, R., Mukherjee, S., R{\"u}hle, V.,
  Lakshmanan, L. V.~S., and Awadallah, A.~H.
\newblock Hybrid {LLM}: Cost-efficient and quality-aware query routing.
\newblock In \emph{International Conference on Learning Representations
  (ICLR)}, pp.\  41348--41366, 2024.
\newblock URL
  \url{https://proceedings.iclr.cc/paper_files/paper/2024/hash/b47d93c99fa22ac0b377578af0a1f63a-Abstract-Conference.html}.

\bibitem[Dong et~al.(2025)Dong, Tang, Liu, Li, Chu, and Li]{quantagent}
Dong, P., Tang, Z., Liu, X., Li, L., Chu, X., and Li, B.
\newblock Can compressed {LLMs} truly act? {An} empirical evaluation of agentic
  capabilities in {LLM} compression.
\newblock In \emph{Proceedings of the 42nd International Conference on Machine
  Learning (ICML)}, volume 267 of \emph{Proceedings of Machine Learning
  Research}, pp.\  14169--14202. PMLR, 2025.
\newblock URL \url{https://proceedings.mlr.press/v267/dong25k.html}.

\bibitem[Fei et~al.(2025)Fei, Ji, Wang, Shi, Gong, and Qiu]{embodiedr1}
Fei, Z., Ji, L., Wang, S., Shi, J., Gong, J., and Qiu, X.
\newblock Unleashing embodied task planning ability in {LLMs} via reinforcement
  learning, 2025.
\newblock URL \url{https://arxiv.org/abs/2506.23127}.

\bibitem[Frantar et~al.(2023)Frantar, Ashkboos, Hoefler, and Alistarh]{gptq}
Frantar, E., Ashkboos, S., Hoefler, T., and Alistarh, D.
\newblock {OPTQ}: Accurate quantization for generative pre-trained
  transformers.
\newblock In \emph{International Conference on Learning Representations
  (ICLR)}, 2023.
\newblock Also known as {GPTQ}, arXiv:2210.17323.

\bibitem[Fu et~al.(2025)Fu, Ge, You, Liu, Yuan, Dai, Yan, Yang, and Wang]{r2r}
Fu, T., Ge, Y., You, Y., Liu, E., Yuan, Z., Dai, G., Yan, S., Yang, H., and
  Wang, Y.
\newblock {R2R}: Efficiently navigating divergent reasoning paths with
  small-large model token routing.
\newblock In \emph{Advances in Neural Information Processing Systems
  (NeurIPS)}, volume~38, pp.\  124108--124145. Curran Associates, Inc., 2025.
\newblock \doi{10.52202/085713-4135}.

\bibitem[Gunter et~al.(2024)Gunter, Wang, Wang, Pang, Narayanan, Zhang, Zhang,
  Chen, Chiu, Qiu, Gopinath, Yap, Yin, Nan, Weers, Yin, Huang, Wang, Lu,
  Peebles, Ye, Lee, Du, Chen, Keunebroek, Wiseman, Evans, Lei, Rathod, Kong,
  Du, Li, Wang, Gao, Ahmed, Xu, Lu, Rashid, Jose, Doane, Bencomo, Vanderby,
  Hansen, Jain, Anupama, Kamal, Wu, Brum, Maalouf, Erdenebileg, Dulhanty,
  Parilla, Moritz, Kang, Jimenez, Ladd, Shi, Bai, Chu, Hohman, Kotek, Coleman,
  Li, Bigham, Cao, Lai, Cheung, Shan, Zhou, Li, Qin, Singh, Vega, Zou, Heckman,
  Gardiner, Bowler, Cordell, Cao, Hay, Shahdadpuri, Godwin, Dighe, Rachapudi,
  Tantawi, Frigg, Davarnia, Shah, Guha, Sirovica, Ma, Ma, Wang, Kim, Jayaram,
  Shankar, Paidi, Kumar, Wang, Zheng, Cheng, Shrager, Ye, Tanaka, Guo, Meng,
  Luo, Ouyang, Aygar, Wan, Walkingshaw, Lin, Farooq, Ramerth, Reed, Bartels,
  Chaney, Riazati, Yang, Feldman, Hochstrasser, Seguin, Belousova, Pelemans,
  Yang, Vahid, Cao, Najibi, Zuliani, Horton, Cho, Bhendawade, Dong, Maj,
  Agrawal, Shan, Fu, Poston, Xu, Liu, Rao, Heeramun, Merth, Rayala, Cui,
  Sridhar, Zhang, Zhang, Wu, Zhou, Liu, Zhao, Xia, Ren, and Ren]{apple-afm}
Gunter, T., Wang, Z., Wang, C., Pang, R., Narayanan, A., Zhang, A., Zhang, B.,
  Chen, C., Chiu, C.-C., Qiu, D., Gopinath, D., Yap, D.~A., Yin, D., Nan, F.,
  Weers, F., Yin, G., Huang, H., Wang, J., Lu, J., Peebles, J., Ye, K., Lee,
  M., Du, N., Chen, Q., Keunebroek, Q., Wiseman, S., Evans, S., Lei, T.,
  Rathod, V., Kong, X., Du, X., Li, Y., Wang, Y., Gao, Y., Ahmed, Z., Xu, Z.,
  Lu, Z., Rashid, A., Jose, A.~M., Doane, A., Bencomo, A., Vanderby, A.,
  Hansen, A., Jain, A., Anupama, A.~M., Kamal, A., Wu, B., Brum, C., Maalouf,
  C., Erdenebileg, C., Dulhanty, C., Parilla, D., Moritz, D., Kang, D.,
  Jimenez, E., Ladd, E., Shi, F., Bai, F., Chu, F., Hohman, F., Kotek, H.,
  Coleman, H.~G., Li, J., Bigham, J., Cao, J., Lai, J., Cheung, J., Shan, J.,
  Zhou, J., Li, J., Qin, J., Singh, K., Vega, K., Zou, K., Heckman, L.,
  Gardiner, L., Bowler, M., Cordell, M., Cao, M., Hay, N., Shahdadpuri, N.,
  Godwin, O., Dighe, P., Rachapudi, P., Tantawi, R., Frigg, R., Davarnia, S.,
  Shah, S., Guha, S., Sirovica, S., Ma, S., Ma, S., Wang, S., Kim, S., Jayaram,
  S., Shankar, V., Paidi, V., Kumar, V., Wang, X., Zheng, X., Cheng, W.,
  Shrager, Y., Ye, Y., Tanaka, Y., Guo, Y., Meng, Y., Luo, Z.~T., Ouyang, Z.,
  Aygar, A., Wan, A., Walkingshaw, A., Lin, A., Farooq, A., Ramerth, B., Reed,
  C., Bartels, C., Chaney, C., Riazati, D., Yang, E.~L., Feldman, E.,
  Hochstrasser, G., Seguin, G., Belousova, I., Pelemans, J., Yang, K., Vahid,
  K.~A., Cao, L., Najibi, M., Zuliani, M., Horton, M., Cho, M., Bhendawade, N.,
  Dong, P., Maj, P., Agrawal, P., Shan, Q., Fu, Q., Poston, R., Xu, S., Liu,
  S., Rao, S., Heeramun, T., Merth, T., Rayala, U., Cui, V., Sridhar, V.~R.,
  Zhang, W., Zhang, W., Wu, W., Zhou, X., Liu, X., Zhao, Y., Xia, Y., Ren, Z.,
  and Ren, Z.
\newblock Apple intelligence foundation language models, 2024.
\newblock URL \url{https://arxiv.org/abs/2407.21075}.

\bibitem[He et~al.(2025)He, Ding, Cai, Zhang, Xiong, Sun, Qin, and
  Liu]{selfroute}
He, Y., Ding, X., Cai, B., Zhang, Y., Xiong, K., Sun, Z., Qin, B., and Liu, T.
\newblock {Self-Route}: Automatic mode switching via capability estimation for
  efficient reasoning, 2025.
\newblock URL \url{https://arxiv.org/abs/2505.20664}.

\bibitem[Jiang \& Huang(2026)Jiang and Huang]{openphone}
Jiang, Y. and Huang, C.
\newblock {OpenPhone}: Mobile agentic foundation models.
\newblock In \emph{Findings of the Association for Computational Linguistics:
  ACL 2026}, pp.\  30362--30380. Association for Computational Linguistics,
  2026.
\newblock \doi{10.18653/v1/2026.findings-acl.1518}.

\bibitem[Jin et~al.(2024)Jin, Du, Huang, Liu, Luan, Wang, and
  Xiong]{jin-etal-2024-comprehensive}
Jin, R., Du, J., Huang, W., Liu, W., Luan, J., Wang, B., and Xiong, D.
\newblock A comprehensive evaluation of quantization strategies for large
  language models.
\newblock In Ku, L.-W., Martins, A., and Srikumar, V. (eds.), \emph{Findings of
  the Association for Computational Linguistics: ACL 2024}, pp.\  12186--12215,
  Bangkok, Thailand, August 2024. Association for Computational Linguistics.
\newblock \doi{10.18653/v1/2024.findings-acl.726}.
\newblock URL \url{https://aclanthology.org/2024.findings-acl.726/}.

\bibitem[Kapoor et~al.(2026)Kapoor, Gupta, Chen, Beniwal, Huang, and
  Kumar]{trim}
Kapoor, V., Gupta, A., Chen, H., Beniwal, A., Huang, J., and Kumar, A.
\newblock {TRIM}: Hybrid inference via targeted stepwise routing in multi-step
  reasoning tasks.
\newblock In \emph{International Conference on Learning Representations
  (ICLR)}, 2026.

\bibitem[Kwon et~al.(2025)Kwon, Seo, Lee, and Park]{dpllm}
Kwon, S., Seo, S.~H., Lee, J.~W., and Park, Y.
\newblock {DP-LLM}: Runtime model adaptation with dynamic layer-wise precision
  assignment.
\newblock In \emph{Advances in Neural Information Processing Systems
  (NeurIPS)}, volume~38, pp.\  144921--144945. Curran Associates, Inc., 2025.
\newblock \doi{10.52202/085713-4849}.

\bibitem[Kwon et~al.(2023)Kwon, Li, Zhuang, Sheng, Zheng, Yu, Gonzalez, Zhang,
  and Stoica]{vllm}
Kwon, W., Li, Z., Zhuang, S., Sheng, Y., Zheng, L., Yu, C.~H., Gonzalez, J.~E.,
  Zhang, H., and Stoica, I.
\newblock Efficient memory management for large language model serving with
  {PagedAttention}.
\newblock In \emph{Proceedings of the 29th Symposium on Operating Systems
  Principles (SOSP)}, pp.\  611--626. ACM, 2023.
\newblock \doi{10.1145/3600006.3613165}.

\bibitem[Lee et~al.(2026)Lee, Kim, Koh, Yang, and Jung]{steer}
Lee, S., Kim, D., Koh, H., Yang, N., and Jung, K.
\newblock Confidence-guided stepwise model routing for cost-efficient
  reasoning.
\newblock In \emph{Proceedings of the AAAI Conference on Artificial
  Intelligence (AAAI)}, volume~40, pp.\  31483--31491, 2026.
\newblock \doi{10.1609/aaai.v40i37.40413}.

\bibitem[Li et~al.(2026)Li, Zhang, Cui, Mu, Zhang, Zhang, Jia, and
  Hu]{rsirouter}
Li, H., Zhang, H., Cui, Z., Mu, C., Zhang, Y., Zhang, B., Jia, D., and Hu, S.
\newblock {RSI-Router}: Evolving subtask-level {LLM} routing and skills for
  cost-efficient agents, 2026.
\newblock URL \url{https://arxiv.org/abs/2609.34712}.

\bibitem[Liang et~al.(2026)Liang, Yu, Nasiri, and Garraghan]{geomdiv}
Liang, J., Yu, Z., Nasiri, H., and Garraghan, P.
\newblock Geometry of divergence: Tracking hidden-state trajectories for
  adaptive multi-turn reasoning, 2026.
\newblock URL \url{https://arxiv.org/abs/2608.30650}.

\bibitem[Lin et~al.(2024)Lin, Tang, Tang, Yang, Chen, Wang, Xiao, Dang, Gan,
  and Han]{awq}
Lin, J., Tang, J., Tang, H., Yang, S., Chen, W.-M., Wang, W.-C., Xiao, G.,
  Dang, X., Gan, C., and Han, S.
\newblock {AWQ}: Activation-aware weight quantization for on-device {LLM}
  compression and acceleration.
\newblock In \emph{Proceedings of Machine Learning and Systems (MLSys)},
  volume~6, pp.\  87--100, 2024.
\newblock URL
  \url{https://proceedings.mlsys.org/paper_files/paper/2024/hash/42a452cbafa9dd64e9ba4aa95cc1ef21-Abstract-Conference.html}.

\bibitem[Long et~al.(2026)Long, Wei, Wu, Wei, Huang, and Du]{edgeagent}
Long, Y., Wei, Y., Wu, K., Wei, J., Huang, D., and Du, J.
\newblock {EdgeAgent}: Orchestrating on-device {LLM} inference for end-user
  multi-agent systems on {CPU-GPU} unified memory architectures, 2026.
\newblock URL \url{https://arxiv.org/abs/2610.03394}.

\bibitem[Lu et~al.(2026)Lu, Chen, Fang, Ma, and Wang]{mixquant}
Lu, H., Chen, Z., Fang, G., Ma, X., and Wang, X.
\newblock {Mix-Quant}: Quantized prefilling, precise decoding for agentic
  {LLMs}, 2026.
\newblock URL \url{https://arxiv.org/abs/2605.20315}.

\bibitem[Ma et~al.(2018)Ma, Zhao, Huang, Wang, Hu, Zhu, and Gai]{esmm}
Ma, X., Zhao, L., Huang, G., Wang, Z., Hu, Z., Zhu, X., and Gai, K.
\newblock Entire space multi-task model: An effective approach for estimating
  post-click conversion rate.
\newblock In \emph{Proceedings of the 41st International ACM SIGIR Conference
  on Research and Development in Information Retrieval (SIGIR)}, pp.\
  1137--1140, 2018.
\newblock \doi{10.1145/3209978.3210104}.

\bibitem[Nair et~al.(2025)Nair, Datta, Dean, Jain, and Kusupati]{matquant}
Nair, P.~A., Datta, P., Dean, J., Jain, P., and Kusupati, A.
\newblock Matryoshka quantization.
\newblock In \emph{Proceedings of the 42nd International Conference on Machine
  Learning (ICML)}, volume 267 of \emph{Proceedings of Machine Learning
  Research}, pp.\  45484--45506. PMLR, 2025.
\newblock URL \url{https://proceedings.mlr.press/v267/nair25a.html}.

\bibitem[Ong et~al.(2025)Ong, Almahairi, Wu, Chiang, Wu, Gonzalez, Kadous, and
  Stoica]{routellm}
Ong, I., Almahairi, A., Wu, V., Chiang, W.-L., Wu, T., Gonzalez, J.~E., Kadous,
  M.~W., and Stoica, I.
\newblock {RouteLLM}: Learning to route {LLMs} from preference data.
\newblock In \emph{International Conference on Learning Representations
  (ICLR)}, pp.\  34433--34448, 2025.
\newblock URL
  \url{https://proceedings.iclr.cc/paper_files/paper/2025/hash/5503a7c69d48a2f86fc00b3dc09de686-Abstract-Conference.html}.

\bibitem[Park et~al.(2026)Park, Bae, Kwon, Kim, Kwon, and Lee]{anybcq}
Park, G., Bae, J., Kwon, B., Kim, B., Kwon, S.~J., and Lee, D.
\newblock {AnyBCQ}: Hardware efficient flexible binary-coded quantization for
  multi-precision {LLMs}.
\newblock In \emph{International Conference on Learning Representations
  (ICLR)}, pp.\  125106--125122, 2026.
\newblock URL
  \url{https://proceedings.iclr.cc/paper_files/paper/2026/hash/ca9eaef07eca2a50fc626cb929617b1c-Abstract-Conference.html}.

\bibitem[Park et~al.(2024)Park, Hyun, Cho, Sim, and Lee]{anyprecision}
Park, Y., Hyun, J., Cho, S., Sim, B., and Lee, J.~W.
\newblock Any-precision {LLM}: Low-cost deployment of multiple, different-sized
  {LLMs}.
\newblock In \emph{Proceedings of the 41st International Conference on Machine
  Learning (ICML)}, volume 235 of \emph{Proceedings of Machine Learning
  Research}, pp.\  39682--39701. PMLR, 2024.
\newblock URL \url{https://proceedings.mlr.press/v235/park24e.html}.

\bibitem[Piatrashyn et~al.(2026)Piatrashyn, Kotelevskii, Grishchenkov, Glazkov,
  Nasonov, Makarov, Baldwin, Nakov, Vashurin, and Panov]{redact}
Piatrashyn, D., Kotelevskii, N., Grishchenkov, K., Glazkov, N., Nasonov, I.,
  Makarov, I., Baldwin, T., Nakov, P., Vashurin, R., and Panov, M.
\newblock {ReDAct}: Uncertainty-aware deferral for {LLM} agents, 2026.
\newblock URL \url{https://arxiv.org/abs/2604.07036}.

\bibitem[Ross et~al.(2011)Ross, Gordon, and Bagnell]{dagger}
Ross, S., Gordon, G.~J., and Bagnell, J.~A.
\newblock A reduction of imitation learning and structured prediction to
  no-regret online learning.
\newblock In \emph{Proceedings of the Fourteenth International Conference on
  Artificial Intelligence and Statistics (AISTATS)}, volume~15 of
  \emph{Proceedings of Machine Learning Research}, pp.\  627--635. PMLR, 2011.
\newblock URL \url{https://proceedings.mlr.press/v15/ross11a.html}.

\bibitem[Schick et~al.(2023)Schick, Dwivedi-Yu, Dess{\`\i}, Raileanu, Lomeli,
  Hambro, Zettlemoyer, Cancedda, and Scialom]{schick2023toolformer}
Schick, T., Dwivedi-Yu, J., Dess{\`\i}, R., Raileanu, R., Lomeli, M., Hambro,
  E., Zettlemoyer, L., Cancedda, N., and Scialom, T.
\newblock Toolformer: Language models can teach themselves to use tools.
\newblock In \emph{Advances in Neural Information Processing Systems
  (NeurIPS)}, volume~36, pp.\  68539--68551. Curran Associates, Inc., 2023.
\newblock \doi{10.52202/075280-2997}.

\bibitem[Shinn et~al.(2023)Shinn, Cassano, Gopinath, Narasimhan, and
  Yao]{shinn2023reflexion}
Shinn, N., Cassano, F., Gopinath, A., Narasimhan, K., and Yao, S.
\newblock Reflexion: Language agents with verbal reinforcement learning.
\newblock In \emph{Advances in Neural Information Processing Systems
  (NeurIPS)}, volume~36, pp.\  8634--8652. Curran Associates, Inc., 2023.
\newblock \doi{10.52202/075280-0377}.

\bibitem[Shridhar et~al.(2021)Shridhar, Yuan, C{\^o}t{\'e}, Bisk, Trischler,
  and Hausknecht]{alfworld}
Shridhar, M., Yuan, X., C{\^o}t{\'e}, M.-A., Bisk, Y., Trischler, A., and
  Hausknecht, M.
\newblock {ALFWorld}: Aligning text and embodied environments for interactive
  learning.
\newblock In \emph{International Conference on Learning Representations
  (ICLR)}, 2021.
\newblock URL \url{https://openreview.net/forum?id=0IOX0YcCdTn}.

\bibitem[Song et~al.(2024)Song, Mi, Xie, and Chen]{powerinfer}
Song, Y., Mi, Z., Xie, H., and Chen, H.
\newblock {PowerInfer}: Fast large language model serving with a consumer-grade
  {GPU}.
\newblock In \emph{Proceedings of the 30th Symposium on Operating Systems
  Principles (SOSP)}, pp.\  590--606. ACM, 2024.
\newblock \doi{10.1145/3694715.3695964}.

\bibitem[Su et~al.(2026)Su, Zhang, Lan, Wang, Shen, Yang, and
  Cheng]{morphserve}
Su, Z., Zhang, Z., Lan, T., Wang, Z., Shen, H., Yang, J., and Cheng, Y.
\newblock {MorphServe}: Efficient and workload-aware {LLM} serving via runtime
  quantized layer swapping and {KV} cache resizing.
\newblock In \emph{Proceedings of Machine Learning and Systems (MLSys)},
  volume~8, pp.\  1056--1075, 2026.
\newblock URL
  \url{https://proceedings.mlsys.org/paper_files/paper/2026/hash/8144a9d62e506af0fcdeac0e456b2710-Abstract-Conference.html}.

\bibitem[Taha et~al.(2026)Taha, Uddin, and Alam]{agentcompress}
Taha, Z. A.~K., Uddin, M.~M., and Alam, S.
\newblock {AgentCompress}: Task-aware compression for affordable large language
  model agents, 2026.
\newblock URL \url{https://arxiv.org/abs/2601.05191}.

\bibitem[Varshney et~al.(2026)Varshney, Surla, Xu, Krishnan, Jeblick, Austin,
  Vaidya, and Onofrio]{prefillrouter}
Varshney, T., Surla, A., Xu, M., Krishnan, G.~V., Jeblick, M., Austin, D.,
  Vaidya, N., and Onofrio, D.
\newblock {LLM} router: Rethinking routing with prefill activations, 2026.
\newblock URL \url{https://arxiv.org/abs/2603.20895}.

\bibitem[Wang et~al.(2024)Wang, Xie, Jiang, Mandlekar, Xiao, Zhu, Fan, and
  Anandkumar]{DBLP:journals/tmlr/WangX0MXZFA24}
Wang, G., Xie, Y., Jiang, Y., Mandlekar, A., Xiao, C., Zhu, Y., Fan, L., and
  Anandkumar, A.
\newblock Voyager: An open-ended embodied agent with large language models.
\newblock \emph{Transactions on Machine Learning Research}, 2024.
\newblock URL \url{https://openreview.net/forum?id=ehfRiF0R3a}.

\bibitem[Wang et~al.(2022)Wang, Jansen, C{\^o}t{\'e}, and
  Ammanabrolu]{scienceworld}
Wang, R., Jansen, P., C{\^o}t{\'e}, M.-A., and Ammanabrolu, P.
\newblock {ScienceWorld}: Is your agent smarter than a 5th grader?
\newblock In \emph{Proceedings of the 2022 Conference on Empirical Methods in
  Natural Language Processing (EMNLP)}, pp.\  11279--11298. Association for
  Computational Linguistics, 2022.
\newblock \doi{10.18653/v1/2022.emnlp-main.775}.

\bibitem[Wei et~al.(2025)Wei, Zhang, Li, Chen, Guan, Qu, Li, Chen, and
  Luo]{agentxpu}
Wei, X., Zhang, J., Li, H., Chen, J., Guan, H., Qu, R., Li, M., Chen, X., and
  Luo, G.
\newblock {Agent.xpu}: Efficient scheduling of agentic {LLM} workloads on
  heterogeneous {SoC}, 2025.
\newblock URL \url{https://arxiv.org/abs/2506.24045}.

\bibitem[Yang et~al.(2025{\natexlab{a}})Yang, Li, Yang, Zhang, Hui, Zheng, Yu,
  Gao, Huang, Lv, Zheng, Liu, Zhou, Huang, Hu, Ge, Wei, Lin, Tang, Yang, Tu,
  Zhang, Yang, Yang, Zhou, Zhou, Lin, Dang, Bao, Yang, Yu, Deng, Li, Xue, Li,
  Zhang, Wang, Zhu, Men, Gao, Liu, Luo, Li, Tang, Yin, Ren, Wang, Zhang, Ren,
  Fan, Su, Zhang, Zhang, Wan, Liu, Wang, Cui, Zhang, Zhou, and Qiu]{qwen3}
Yang, A., Li, A., Yang, B., Zhang, B., Hui, B., Zheng, B., Yu, B., Gao, C.,
  Huang, C., Lv, C., Zheng, C., Liu, D., Zhou, F., Huang, F., Hu, F., Ge, H.,
  Wei, H., Lin, H., Tang, J., Yang, J., Tu, J., Zhang, J., Yang, J., Yang, J.,
  Zhou, J., Zhou, J., Lin, J., Dang, K., Bao, K., Yang, K., Yu, L., Deng, L.,
  Li, M., Xue, M., Li, M., Zhang, P., Wang, P., Zhu, Q., Men, R., Gao, R., Liu,
  S., Luo, S., Li, T., Tang, T., Yin, W., Ren, X., Wang, X., Zhang, X., Ren,
  X., Fan, Y., Su, Y., Zhang, Y., Zhang, Y., Wan, Y., Liu, Y., Wang, Z., Cui,
  Z., Zhang, Z., Zhou, Z., and Qiu, Z.
\newblock {Qwen3} technical report, 2025{\natexlab{a}}.
\newblock URL \url{https://arxiv.org/abs/2505.09388}.

\bibitem[Yang et~al.(2024)Yang, Jimenez, Wettig, Lieret, Yao, Narasimhan, and
  Press]{DBLP:conf/nips/YangJWLYNP24}
Yang, J., Jimenez, C.~E., Wettig, A., Lieret, K., Yao, S., Narasimhan, K., and
  Press, O.
\newblock Swe-agent: Agent-computer interfaces enable automated software
  engineering.
\newblock In Globersons, A., Mackey, L., Belgrave, D., Fan, A., Paquet, U.,
  Tomczak, J.~M., and Zhang, C. (eds.), \emph{Advances in Neural Information
  Processing Systems 37: Annual Conference on Neural Information Processing
  Systems 2024, NeurIPS 2024, Vancouver, BC, Canada, December 10 - 15, 2024},
  pp.\  50528--50652, 2024.
\newblock \doi{10.52202/079017-1601}.

\bibitem[Yang et~al.(2025{\natexlab{b}})Yang, Dou, Feng, Huang, Nguyen, You,
  Attia, Yang, Feng, Zhang, Ramrakhya, Jia, Nichols, Toshev, Yang, and
  Gan]{ferretuilite}
Yang, Z., Dou, Z.-Y., Feng, D., Huang, F., Nguyen, A., You, K., Attia, O.,
  Yang, Y., Feng, M., Zhang, H., Ramrakhya, R., Jia, C., Nichols, J., Toshev,
  A., Yang, Y., and Gan, Z.
\newblock {Ferret-UI Lite}: Lessons from building small on-device {GUI} agents,
  2025{\natexlab{b}}.
\newblock URL \url{https://arxiv.org/abs/2509.26539}.

\bibitem[Yao et~al.(2023)Yao, Zhao, Yu, Du, Shafran, Narasimhan, and
  Cao]{react}
Yao, S., Zhao, J., Yu, D., Du, N., Shafran, I., Narasimhan, K., and Cao, Y.
\newblock {ReAct}: Synergizing reasoning and acting in language models.
\newblock In \emph{International Conference on Learning Representations
  (ICLR)}, 2023.
\newblock URL \url{https://openreview.net/forum?id=WE_vluYUL-X}.

\bibitem[Zhang et~al.(2025)Zhang, Feng, and You]{routerr1}
Zhang, H., Feng, T., and You, J.
\newblock {Router-R1}: Teaching {LLMs} multi-round routing and aggregation via
  reinforcement learning.
\newblock In \emph{Advances in Neural Information Processing Systems
  (NeurIPS)}, volume~38, pp.\  141233--141265. Curran Associates, Inc., 2025.
\newblock \doi{10.52202/085713-4719}.

\bibitem[Zhang et~al.(2026{\natexlab{a}})Zhang, Hu, Lin, Fan, Xie, Fang, Yang,
  Zhu, Chen, Lv, and Chen]{hera}
Zhang, Y., Hu, M., Lin, Z., Fan, X., Xie, F., Fang, Z., Yang, J., Zhu, W.,
  Chen, Z., Lv, C., and Chen, Z.
\newblock {Hera}: Learning long-horizon coordination for device-cloud
  collaborative {LLM} agents, 2026{\natexlab{a}}.
\newblock URL \url{https://arxiv.org/abs/2605.24598}.

\bibitem[Zhang et~al.(2026{\natexlab{b}})Zhang, Li, Wang, Feng, Yang, Wang,
  Zhang, Bai, and Hu]{mtrouter}
Zhang, Y., Li, H., Wang, Z., Feng, S., Yang, X., Wang, D., Zhang, B., Bai, L.,
  and Hu, S.
\newblock {MTRouter}: Cost-aware multi-turn {LLM} routing with history--model
  joint embeddings.
\newblock In \emph{Proceedings of the 64th Annual Meeting of the Association
  for Computational Linguistics (Volume 1: Long Papers) (ACL)}, pp.\
  44206--44226. Association for Computational Linguistics, 2026{\natexlab{b}}.
\newblock \doi{10.18653/v1/2026.acl-long.2045}.

\bibitem[Zhao et~al.(2025)Zhao, Lu, Wang, Kong, and Wu]{qspec}
Zhao, J., Lu, W., Wang, S., Kong, L., and Wu, C.
\newblock {QSpec}: Speculative decoding with complementary quantization
  schemes.
\newblock In \emph{Proceedings of the 2025 Conference on Empirical Methods in
  Natural Language Processing (EMNLP)}, pp.\  4779--4795. Association for
  Computational Linguistics, 2025.
\newblock \doi{10.18653/v1/2025.emnlp-main.240}.

\bibitem[Zhuang et~al.(2025)Zhuang, Wu, Wen, Li, Jiao, and
  Ramchandran]{embedllm}
Zhuang, R., Wu, T., Wen, Z., Li, A., Jiao, J., and Ramchandran, K.
\newblock {EmbedLLM}: Learning compact representations of large language
  models.
\newblock In \emph{International Conference on Learning Representations
  (ICLR)}, pp.\  76913--76926, 2025.
\newblock URL
  \url{https://proceedings.iclr.cc/paper_files/paper/2025/hash/bf5e4b85d203481d6e37bd32d9600162-Abstract-Conference.html}.

\end{thebibliography}
\bibliographystyle{mlsys2025}

\clearpage
\appendix
\counterwithin{table}{section}
\counterwithin{figure}{section}
\counterwithin{equation}{section}
\section{Benchmark Protocols}
\label{app:protocols}

\begin{table}[h]
\centering
\caption{Benchmarks, protocols and per-step workload. Workload rows are measured on the FP16 agents' test trajectories; prompt and response lengths are in tokens, averaged over all steps.}
\label{tab:workload}
\footnotesize
\setlength{\tabcolsep}{3pt}
\resizebox{\columnwidth}{!}{%
\begin{tabular}{lccc}
\toprule
 & ALFWorld & SciWorld & TextWorld \\
\midrule
Agent model          & Qwen3-4B & ER1-7B & Qwen3-4B \\
Prompt format        & ReAct, 2-shot & Thought/Action & ReAct, 1-shot \\
Max tokens / calls   & 100 / 50 & 200 / 30 & 100 / 50 \\
Context limit        & 8192 & 4096 & 8192 \\
Test tasks           & 140 + 134 & 211 & 100 \\
\midrule
Steps per episode    & 22.2 & 13.9 & 32.9 \\
Prompt length (max)  & 2011 (3864) & 1775 (3886) & 2028 (5822) \\
Response length      & 15.3 & 35.3 & 11.1 \\
\bottomrule
\end{tabular}}
\end{table}

Table~\ref{tab:workload} summarizes the benchmarks, protocols and per-step workload.

\para{Experimental setup and measurement details.}
\textbf{Agent protocols.}
All agents retain the full interaction history and use greedy decoding.
ALFWorld uses two-shot ReAct prompting~\cite{react}, TextWorld uses a one-game demonstration (Appendix~\ref{app:textworld}), and ScienceWorld uses ER1-7B from Embodied Planner-R1~\cite{embodiedr1}.
ALFWorld success is evaluated on the seen and unseen test splits, while ScienceWorld success requires a final score of 100.

\textbf{Quantization configurations.}
The top tier achieves success rates comparable to FP16: 88.3\% versus 87.2\% for W8 on ALFWorld, and 71.1\% versus 71.6\% for W4 on ScienceWorld.
For independent PTQ, GPTQ uses group size 128, asymmetric quantization, and 128 C4 sequences of 2048 tokens.

\textbf{Latency measurement and controls.}
We measure TPOT at a context length of 2,048 tokens, close to the mean per-call context length (2.0k on ALFWorld, 1.8k on ScienceWorld, and 2.0k on TextWorld), using our Any-Precision engine and vLLM's GPTQ kernels~\cite{vllm}.
Decode time sums the output-token count multiplied by the decoding tier's TPOT across calls, with all costs aggregated per task to account for changes in trajectory length.
The random baseline matches \sys{}'s W3 token share within 1.5 percentage points; for example, on ALFWorld with GPTQ, $p=0.78$ yields 78.7\% versus \sys{}'s 78.0\%.
All methods within a precision setting use the same engine.

\para{ALFWorld.}
Qwen3-4B receives the two in-context examples of the original ReAct prompt for the task type and the full history of the episode.
Each call generates at most 100 tokens greedily and stops at the first newline; thought lines are passed to the environment, which answers them with ``OK.''.
An episode ends on success or after 50 calls.

\para{ScienceWorld.}
ER1-7B uses its original instruction and the full conversation history.
Responses have at most 200 new tokens, and the text after \texttt{Action:} is executed.
An invalid response consumes a step.
An episode ends on completion or after 30 calls; success means a final score of 100.

\para{Router thresholds.}
The main results use $\tau = 0.997$ on ScienceWorld, $0.95$ on TextWorld, and $0.96$ and $0.95$ on ALFWorld with Any-Precision and GPTQ.

\section{Label Details}
\label{app:labels}

Table~\ref{tab:label-stats} reports the states collected under the round-0 router, with a counterfactual branch at every divergent state (\S\ref{sec:sup-states}).

\begin{table}[h]
\centering
\caption{Round-1 counterfactual labels (training split). Critical: the low-tier branch fails while the top-tier branch succeeds.}
\label{tab:label-stats}
\scriptsize
\setlength{\tabcolsep}{2pt}
\begin{tabular}{lccc}
\toprule
 & States & Divergent & Critical \\
\midrule
ALFWorld GPTQ $\{\text{W3}, \text{FP16}\}$ & 13{,}450 & 2{,}049 & 143 (1.1\%) \\
ALFWorld AP $\{\text{W3}, \text{W8}\}$     & 13{,}232 & 2{,}162 & 134 (1.0\%) \\
ScienceWorld AP $\{\text{W3}, \text{W4}\}$ & 13{,}358 & 1{,}747 & 80 (0.6\%) \\
TextWorld AP $\{\text{W3}, \text{W8}\}$    & 18{,}032 & 3{,}016 & 408 (2.3\%) \\
\bottomrule
\end{tabular}
\end{table}

\section{Additional Ablations}
\label{app:ablations}

\para{Why the criticality labels matter.}
The round-0 router predicts only whether the two precisions will produce different actions, treating every predicted divergence as critical ($p_K \equiv 1$ in Equation~\ref{eq:esmm}).
Consequently, it routes many harmless divergences to FP16, limiting W3 decoding to 34--56\% of tokens and achieving at most $1.28\times$ speedup on ALFWorld with GPTQ (Table~\ref{tab:round0}).
Lowering the routing threshold increases W3 usage but reduces task success (84.3\% at $\tau=0.5$).
With the criticality head trained on round-1 counterfactual branches, \sys{} decodes 78\% of tokens at W3 and achieves $1.71\times$ speedup with a higher success rate of 86.9\%.

\begin{table}[t]
\centering
\caption{Divergence-only routing versus \sys{} on ALFWorld (GPTQ $\{\text{W3},\text{FP16}\}$, FP16 prefill). Speedup is the mean decode time per task of FP16 divided by that of each router; W3 is the fraction of tokens decoded at W3. Thresholds apply to different scores: $1-p_D$ for Round 0 and $1-p_Dp_K$ for \sys{}.}
\label{tab:round0}
\small
\setlength{\tabcolsep}{5pt}
\begin{tabular}{lcccc}
\toprule
Router & $\tau$ & Succ. & Speedup & W3 \\
\midrule
FP16 & -- & 87.2 & 1.00$\times$ & 0\% \\
\midrule
Round 0 & 0.5 & 84.3 & 1.27$\times$ & 56\% \\
 & 0.7 & 86.1 & 1.28$\times$ & 46\% \\
 & 0.9 & 86.5 & 1.22$\times$ & 34\% \\
\midrule
\rowcolor{oursrow} \sys{} & 0.95 & 86.9 & 1.71$\times$ & 78\% \\
\bottomrule
\end{tabular}
\end{table}

\para{Threshold sweep.}
Table~\ref{tab:tau-sweep} reports five thresholds around the main operating point for ALFWorld with GPTQ and ScienceWorld, with rows ordered by W3 token share.
No heuristic baseline achieves both higher success and lower decode time than \sys{} at any of these thresholds.
On ALFWorld, success ranges from 83.9\% to 86.9\%, while decode time increases with the threshold.
On ScienceWorld, W3 reduces per-token decode time by only 18\% relative to W4, resulting in a narrow range of decode times across thresholds (3.39--3.52\,s).
Across these thresholds, \sys{} also achieves higher success than all heuristic baselines (65.9--68.2\% versus at most 63.8\%).

\begin{table}[h]
\centering
\caption{Results of \sys{} at different thresholds $\tau$ on the test tasks, ordered by W3 token share. W3: W3 token share (\%). Succ.: success rate (\%). Dec.: decode seconds per task. Bold: the threshold of the main results. Heuristic baselines are those of Table~\ref{tab:gptq} (ALFWorld) and Table~\ref{tab:main} (ScienceWorld).}
\label{tab:tau-sweep}
\footnotesize
\setlength{\tabcolsep}{2.5pt}
\resizebox{\columnwidth}{!}{%
\begin{tabular}{lccc@{\hspace{8pt}}lccc}
\toprule
\multicolumn{4}{c}{ALFWorld, GPTQ $\{\text{W3}, \text{FP16}\}$} & \multicolumn{4}{c}{ScienceWorld, AP $\{\text{W3}, \text{W4}\}$} \\
\cmidrule(lr){1-4}\cmidrule(lr){5-8}
$\tau$ & W3 & Succ. & Dec. & $\tau$ & W3 & Succ. & Dec. \\
\midrule
0.90 & 92 & 84.7 & 2.49 & 0.99 & 88 & 65.9 & 3.52 \\
0.93 & 84 & 83.9 & 2.75 & 0.995 & 83 & 67.8 & 3.47 \\
\textbf{0.95} & \textbf{78} & \textbf{86.9} & \textbf{2.86} & \textbf{0.997} & \textbf{77} & \textbf{68.2} & \textbf{3.39} \\
0.96 & 76 & 86.1 & 3.05 & 0.998 & 73 & 67.8 & 3.41 \\
0.97 & 71 & 86.9 & 3.15 & 0.9995 & 58 & 67.8 & 3.45 \\
\midrule
\multicolumn{4}{l}{\emph{Heuristic baselines}} & & & & \\
Step $k{=}2$ & 75 & 85.0 & 2.99 & Step $k{=}2$ & 84 & 63.0 & 3.64 \\
Step $k{=}4$ & 70 & 85.4 & 3.06 & Step $k{=}4$ & 69 & 63.5 & 3.61 \\
Step $k{=}6$ & 65 & 86.5 & 3.25 & Step $k{=}6$ & 57 & 63.5 & 3.79 \\
Random & 79 & 78.8 & 3.34 & Random & 77 & 63.8 & 3.53 \\
\bottomrule
\end{tabular}}
\end{table}

\section{Implementation and Latency Measurement}
\label{app:latency}
\label{app:impl}

\para{Measurement setup.}
All latencies are measured as wall-clock time on an RTX A6000 GPU without competing GPU processes, after warm-up and with CUDA synchronization at phase boundaries.
Per-task model latency sums prefill, routing, and decoding time across all calls.
Environment execution time is excluded from model latency; it amounts to about 13\%, 4\% and 1\% of model time per task on ALFWorld, ScienceWorld and TextWorld, respectively.
We measure prefill latency and per-tier time per output token (TPOT) at context lengths from 512 to 8k tokens using fixed token workloads.

\para{Prefill and decode latency by tier.}
Prefill latency varies little across precision tiers.
For context lengths from 512 to 8k tokens, it ranges from 54 to 1000\,ms on the 4B model and from 86 to 1250\,ms on the 7B model.
In contrast, decode TPOT decreases at lower bit widths: from 9.3\,ms (W8) to 5.1\,ms (W3) on the 4B model, and from 8.1\,ms (W4) to 6.6\,ms (W3) on the 7B model, at a 2k-token context.
These measurements support \sys{}'s use of top-tier prefill and precision-adaptive decoding, since lower-precision prefill provides limited acceleration with the evaluated kernels and can reduce accuracy under independent PTQ (Table~\ref{tab:gptq}).

\para{Measured versus modeled time.}
We validate the decode-time model of \S\ref{sec:eval-setup} against wall-clock time in a separate timing run on an exclusive GPU, comparing measured and modeled decode time on the trajectories of that run.
On ScienceWorld (Any-Precision $\{\text{W3}, \text{W4}\}$, 211 test tasks), \sys{} decodes $1.21\times$ faster than W4 per episode, against $1.18\times$ modeled.
On ALFWorld with GPTQ (274 test games), \sys{} decodes $1.70\times$ faster than FP16 per episode, against $1.66\times$ modeled.
In both settings the measured decode speedup is within $0.04\times$ of the modeled one.

\para{Larger batches.}
Table~\ref{tab:batch} estimates the decode speedup of \sys{} on the test trajectories from Table~\ref{tab:main}, using per-tier time per output token (TPOT) measured at batch sizes 1--8.
We assume that all active sequences within a batch use the same precision tier at each decoding step; mixed-precision batches would instead require separate sub-batches for each tier.
As batch size increases, weight reuse reduces the relative advantage of low-bit decoding.
At batch size eight, the speedup over the top tier decreases from $1.40\times$ to $1.10\times$ on ALFWorld and from $1.23\times$ to $1.09\times$ on ScienceWorld.
Nevertheless, \sys{} retains a $1.66\times$ decode speedup over FP16 on ScienceWorld at batch size eight.

\begin{table}[h]
\centering
\caption{Decode speedup of \sys{} versus batch size, TPOT measured at each batch size.}
\label{tab:batch}
\small
\setlength{\tabcolsep}{4pt}
\begin{tabular}{lcccc}
\toprule
Batch size & 1 & 2 & 4 & 8 \\
\midrule
ALFWorld, over W8   & 1.40$\times$ & 1.31$\times$ & 1.25$\times$ & 1.10$\times$ \\
ALFWorld, over FP16 & 2.09$\times$ & 1.83$\times$ & 1.61$\times$ & 1.24$\times$ \\
ScienceWorld, over W4   & 1.23$\times$ & 1.20$\times$ & 1.13$\times$ & 1.09$\times$ \\
ScienceWorld, over FP16 & 3.25$\times$ & 2.88$\times$ & 2.50$\times$ & 1.66$\times$ \\
\bottomrule
\end{tabular}
\end{table}

\section{Switching Cost on Agent Trajectories}
\label{app:switch-traj}
\begin{figure*}[t]
\centering
\includegraphics{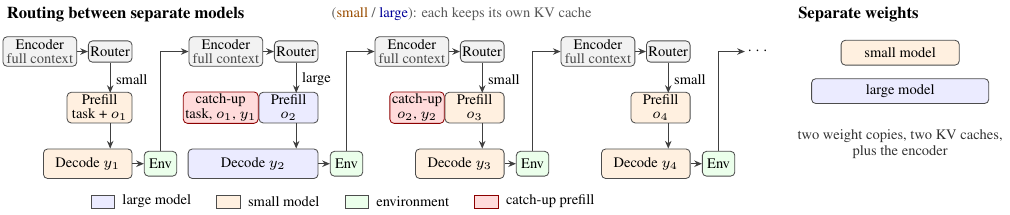}
\caption{Routing between separate models, for contrast with \sys{} (Figure~\ref{fig:framework}a): an external encoder re-reads the context at every step, and a model that was idle in the previous step first prefills the history it missed.}
\label{fig:separate}
\end{figure*}

Figure~\ref{fig:switch-traj} replays the test episodes of \sys{} in Table~\ref{tab:main} and Table~\ref{tab:gptq} (ALFWorld with Any-Precision and GPTQ, ScienceWorld, TextWorld) under four deployments that follow \sys{}'s routing decisions.
The prefill time of a call with $n$ new tokens and a prompt of $L$ tokens is $a + b\,n + c\,n\,(L - n/2)$, fitted by least squares to the measured top-tier prefills of these episodes ($R^2$ = 0.71--0.96); both models use the same function, since prefill takes about the same time at every tier (Appendix~\ref{app:latency}).
\emph{One model at the top tier} prefills the previous response and the new observation at every call.
\sys{} prefills the tokens missing from the top-tier cache, which adds the responses decoded at the low tier.
\emph{Separate models with one cache each} prefill, at every call, all tokens since the chosen model's own last call.
\emph{Separate models that re-prefill} prefill the whole prompt whenever the chosen model differs from the model of the previous call.
Relative to one model at the top tier, \sys{} adds 43--88\,ms of prefill per episode (2.1--3.8\%), separate models with one cache each add 173--315\,ms (7.5--18\%), and separate models that re-prefill add 450--1928\,ms (31--54\%).

\section{Cross-Step Cache Policies}
\label{app:kv}

Table~\ref{tab:kv} compares \sys{}'s cross-step cache policy (\S\ref{sec:design-kv}) with a mixed-precision KV-cache policy on ScienceWorld.
The mixed policy directly reuses low-tier KV entries across agent steps, whereas \sys{} recomputes the KV entries of low-tier responses at the top tier.
Although 14\% of the reused cache tokens in the mixed policy originate from W3 decoding, this policy reduces prefill latency by only 2\,ms per call (87 to 85\,ms) while decreasing task success by 6.6 percentage points (68.2\% to 61.6\%).

\begin{table}[h]
\centering
\caption{Cross-step KV-cache policies on ScienceWorld (Any-Precision $\{\text{W3},\text{W4}\}$; router and threshold as in Table~\ref{tab:main}). Speedup is episode decode speedup over FP16. Prefill is measured top-tier prefill latency per call. Low-tier reuse is the fraction of reused KV-cache tokens originally computed at W3.}
\label{tab:kv}
\footnotesize
\setlength{\tabcolsep}{3pt}
\begin{tabular}{lccccc}
\toprule
Cache policy & Succ. & Speedup & W3 & \makecell{Prefill\\(ms)} & \makecell{Low-tier\\reuse} \\
\midrule
\sys{} (re-prefill) & 68.2 & 3.25$\times$ & 77\% & 87 & 0\% \\
\makecell[l]{Mixed (keep\\low-tier KV)} & 61.6 & 3.08$\times$ & 76\% & 85 & 14\% \\
\bottomrule
\end{tabular}
\end{table}

\section{TextWorld Cooking}
\label{app:textworld}

\para{Games and protocol.}
We generate cooking games with the \texttt{tw-cooking} generator of TextWorld~\cite{textworld} (open, cook and cut enabled): odd seeds use a two-ingredient recipe over six rooms, and even seeds a three-ingredient recipe in the kitchen alone.
The generator builds every game, including its map, objects, recipe and winning condition, from these parameters and a seed.
Seeds 1001--1600 train the router and seeds 1--100 the test set.
Qwen3-4B receives one demonstration game, generated with a seed outside these sets, and the full history of the episode as raw text (Figure~\ref{fig:tw-prompt}).
Each call generates at most 100 tokens greedily; an episode ends on winning or losing the game or after 50 calls.

\begin{figure}[t]
\begin{Verbatim}[fontsize=\scriptsize,frame=single,breaklines=true,breaksymbolleft={},breakindent=1em,commandchars=\\\{\}]
You are playing a text-based cooking game. Read the cookbook to learn the recipe, gather the ingredients, process them as the directions say, then prepare meal and eat meal. To roast an ingredient type "cook <ingredient> with oven", to fry it "cook <ingredient> with stove", to grill it "cook <ingredient> with BBQ"; to chop, slice or dice it hold the knife and type "chop <ingredient> with knife" (or slice, dice). Use one game command per line; a line starting with "think:" is a note to yourself. Here is an example.
You are hungry! Let's cook a delicious meal. Check the cookbook in the kitchen for the recipe. Once done, enjoy your meal!
-= Corridor =- \textit{[...]}
> think: To solve the task, I need to find the cookbook and read the recipe, then gather and process the ingredients and prepare the meal.
OK.
\textit{[... inventory, go north, go west, go north ...]}
> examine cookbook
\textit{[... recipe: pork chop, red apple; directions ...]}
> think: The recipe needs pork chop, red apple. The directions are: slice the pork chop, roast the pork chop, chop the red apple, roast the red apple, prepare meal. I will collect the ingredients, process them with the right tools, then prepare meal and eat meal.
OK.
\textit{[... 9 commands, each followed by the game's response ...]}
> eat meal
You eat the meal. Not bad. \textit{[...]}
You scored 8 out of a possible 8, in 17 turns.

Here is the task.
\textit{<initial observation of the game>}
> \textit{<command>}
\textit{<observation>}
>
\end{Verbatim}
\caption{TextWorld prompt: a fixed instruction, one demonstration game and the history of the current game. The model completes the text after the last \texttt{>} with one line. Italic brackets mark text shortened here.}
\label{fig:tw-prompt}
\end{figure}

\end{document}